\pdfoutput=1

\documentclass[11pt]{article}

\usepackage[preprint]{acl}

\usepackage{times}
\usepackage{latexsym}

\usepackage[T1]{fontenc}

\usepackage[utf8]{inputenc}

\usepackage{microtype}

\usepackage{inconsolata}

\usepackage{graphicx}
\usepackage{amssymb}
\usepackage{inconsolata}
\usepackage{amsmath,amssymb}
\usepackage{epsfig,graphicx,caption,subcaption}
\usepackage{algpseudocode}
\usepackage[normalem]{ulem}
\usepackage{amsmath}
\usepackage[linesnumbered,algoruled,boxed,noend]{algorithm2e}
\usepackage{xcolor}
\usepackage{color, colortbl}
\usepackage{multirow,booktabs, hhline}
\usepackage{amsmath, bm}
\usepackage[linesnumbered,algoruled,boxed,noend]{algorithm2e}
\usepackage{wrapfig}
\usepackage{verbatimbox}
\usepackage{kantlipsum}
\usepackage{fancyvrb}
\usepackage[htt]{hyphenat}
\usepackage{pifont}
\usepackage{bbding}
\usepackage{multirow}
\usepackage{pifont}
\usepackage{mdframed}
\usepackage{CJK}
\usepackage{tcolorbox}
\tcbuselibrary{breakable}  

\usepackage{dsfont}
\newcommand{\cmark}{\ding{51}}%
\newcommand{\xmark}{\ding{55}}%

\newcommand{\our}{Meta-Reflection\xspace}
\definecolor{Gray}{gray}{0.92}
\definecolor{Gray_impr}{gray}{0.98}
\newcommand{\eg}{\emph{e.g.}\xspace}
\newcommand{\ie}{\emph{i.e.}\xspace}

\title{Meta-Reflection: A Feedback-Free Reflection Learning Framework}

\author{
     \textbf{Yaoke Wang~\textsuperscript{1,2}\footnotemark[1]},
     \textbf{Yun Zhu\textsuperscript{1}},
     \textbf{Xintong Bao\textsuperscript{2}},
     \textbf{Wenqiao Zhang\textsuperscript{1}\footnotemark[2]},
     \textbf{Suyang Dai\textsuperscript{2}},\\
     \textbf{Kehan Chen\textsuperscript{2}\footnotemark[2]},
     \textbf{Wenqiang Li\textsuperscript{2}},
     \textbf{Gang Huang\textsuperscript{2}},
     \textbf{Siliang Tang\textsuperscript{1}},
     \textbf{Yueting Zhuang\textsuperscript{1}}\\
 \textsuperscript{1}Zhejiang University,   
 \textsuperscript{2}Alibaba Group
 \\
\small{\{wangyaoke, zhuyun\_dcd, wenqiaozhang, siliang, yzhuang\}@zju.edu.cn}\\
\small{\{xintong.bxt, kehan.ckh, tengyuan.hg\}@alibaba-inc.com}\\
\small{sydai16@fudan.edu.cn},
\small{liwenqiang2021@gmail.com}
}

\begin{document}
\maketitle
\renewcommand{\thefootnote}{\fnsymbol{footnote}}
\footnotetext[1]{Work done when interning at Alibaba Group.}
\footnotetext[2]{Corresponding Author.}

\begin{abstract}
Despite the remarkable capabilities of large language models (LLMs) in natural language understanding and reasoning, they often display undesirable behaviors, such as generating hallucinations and unfaithful reasoning. 
A prevalent strategy to mitigate these issues is the use of reflection, which refines responses through an iterative process. However, while promising, reflection heavily relies on high-quality external feedback and requires iterative multi-agent inference processes, thus hindering its practical application. 
In this paper, we propose~\emph{\our}, a novel feedback-free reflection mechanism that necessitates only a single inference pass without external feedback.
Motivated by the human ability to remember and retrieve reflections from past experiences when encountering similar problems, \our integrates reflective insights into a codebook, allowing the historical insights to be stored, retrieved, and used to guide LLMs in problem-solving. 
To thoroughly investigate and evaluate the practicality of \our in real-world scenarios, we introduce an industrial e-commerce benchmark named E-commerce Customer Intent Detection (\texttt{ECID}). Extensive experiments conducted on both public datasets and the \texttt{ECID} benchmark highlight the effectiveness and efficiency of our proposed approach.

\end{abstract}
\section{Introduction}\label{sec:intro}

Large Language Models (LLMs)~\cite{achiam2023gpt,yang2024qwen2,dubey2024llama} have demonstrated exceptional proficiency in diverse natural language processing tasks, \eg, general language understanding~\cite{wei2022emergent}, generation~\cite{pu2023chatgpt}, and reasoning~\cite{wei2022chain,yao2024tree}. 
However, recent quantitative analyses revealed that contemporary frontier LLMs frequently exhibit undesirable and inconsistent behaviors, including unfaithful reasoning~\cite{turpin2024language} and the production of seemingly plausible yet inaccurate hallucinations~\cite{rawte2023survey}, especially when applying for intricate tasks. Such flawed outputs significantly undermine trust in LLMs and pose substantial obstacles to their widespread adoption in real-world applications.

The undesirable phenomenon of LLMs is somewhat similar to human problem-solving, \ie, we humans do not always generate the best answer on our first try in complex real-life scenarios. While dealing with complex problems, individuals has the capacity to actively refine their answers through a cycle of trial, inspection and correction~\cite{pan2023automatically}. This capacity called ~\emph{Reflection}, enables us to perform better than machines in high-level reasoning and would be the most precious capacity for modern AI. To simulate this ability, ~\emph{LLMs' Reflection}~\cite{madaan2024self,shinn2023reflexion} is devised to mitigate the flawed outputs of LLMs, which utilizes feedback from external sources (\eg, the environment or other LLMs) to prompt the models to adapt their responses.
\begin{figure}
    \centering
    \includegraphics[width=1\linewidth]{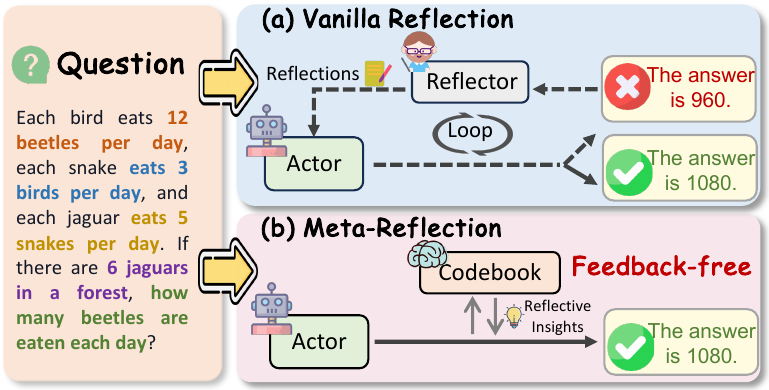}
   \caption{Illustration of different reflection mechanisms. \textbf{(a)}  Vanilla reflection requires multi-agent inference and external feedback. \textbf{(b)}~\our achieves feedback-free reflection in a single inference pass.}
    \label{fig:introduction}
    \vspace{-1.5em}
\end{figure}
This approach, as shown in Figure~\ref{fig:introduction}(a), enables the models to iteratively improve their performance by incorporating new information and adjusting their outputs based on external input, thereby enhancing their accuracy and reliability over time.
Upon reflection, however, contemporary approaches heavily rely on high-quality external feedback or ground-truth golden labels~\cite{huang2024large,dou2024reflection}, which are often unavailable during inference scenarios. 
Besides, reflection typically requires iterative multi-agent inference processes~\cite{du2023improving}, which are resource-intensive. These aforementioned issues significantly constrain the practical deployment of LLMs in real-world scenarios.

In this paper, we propose~\emph{\our}, a novel reflection mechanism that operates without external feedback and requires only a single inference pass. Drawing inspiration from human cognitive processes~\cite{kolodner1992introduction}, where individuals leverage past experiences and reflections to address similar questions without additional trials, we introduce a learnable meta-reflection codebook to store and retrieve reflective insights, as shown in~Figure~\ref{fig:introduction}(b). 
During optimization, reflections are constructed using the vanilla reflection mechanism and integrated into the meta-reflection codebook. At inference, question-specific insights are retrieved from the codebook to guide the LLM in solving problems. This method enables LLMs to produce high-quality responses in a single pass, effectively mimicking how humans utilize prior experiences in analogous situations.
Extensive experiments are conducted with open-source LLMs on diverse benchmarks, including programming, mathematical reasoning, and customer intent detection in E-commerce Intelligent Customer Service (ICS) for industry-specific scenarios. To evaluate our method in the ICS domain, we introduce E-commerce Customer Intent Detection (ECID), a new Chinese dataset designed to identify users' core intents, critical for enhancing service quality. Results across domains validate the efficiency and effectiveness of our approach.
Key contributions of this work include:
\begin{itemize}
    \item  We propose \our, an innovative approach that achieves reflection in a single pass without iterative trials and feedback through well-designed codebook-based storage and retrieval mechanisms.
    \item We present a new dataset for E-commerce Customer Intent Detection (ECID) in the intelligent customer service domain, comprising 1,170 cases from real-world application.
    \item Extensive experiments across various domains and models demonstrate the effectiveness and robustness of our proposed method.
    
\end{itemize}
\section{Related Work}

\subsection{Reflection for Large Language Models}
Large language models (LLMs) \cite{achiam2023gpt, yang2024qwen2, dubey2024llama}, despite their exceptional performance, still exhibit undesired behaviors such as unfaithful reasoning \cite{turpin2024language}, hallucination \cite{rawte2023survey}, and toxic generation \cite{zhang2024efficient}.
\emph{Reflection} techniques~\cite{pan2023automatically, shinn2023reflexion, madaan2024self} address these issues by utilizing feedback to guide LLMs in refining their outputs. For instance, Self-Refine~\cite{madaan2024self} uses a single LLM to generate, critique, and refine outputs, while Reflexion \cite{shinn2023reflexion} employs memory mechanisms and LLM agents to reflect on generations and feedback. \citet{renze2024self} demonstrated the effectiveness of various reflection types across different domains.
Nevertheless, reflection techniques often require high-quality external feedback or golden labels, typically unavailable during deployment \cite{huang2024large, dou2024reflection}, and frequently involve multi-agent inference processes, incurring significant computational costs. While \citet{dou2024reflection} incorporates reflective information through self-training, its implicit incorporation leads to suboptimal results. In this work, we propose~\our, which incorporates reflective information into a learnable codebook, enhancing performance across various tasks.

\subsection{Parameter-Efficient Fine-Tuning (PEFT)}
Parameter-Efficient Fine-Tuning (PEFT) methods enable adaptation of large pretrained models to downstream applications while avoiding the computational costs of full parameter fine-tuning~\cite{hu2023llm}.
These methods can be broadly categorized into two primary approaches: \emph{adapter-based} and \emph{prompt-based} methods.
Adapter-based methods introduce additional trainable parameters to a frozen pretrained model, with notable implementations including LoRA \cite{hu2021lora} and Llama-Adapter \cite{zhang2023llama}. Prompt-based methods transform the discrete optimization of identifying optimal hard prompts into a continuous optimization problem using soft prompts, exemplified by Prefix-Tuning \cite{li2021prefix}, Prompt-Tuning \cite{lester2021power}, and P-Tuning \cite{liu2022ptuning}.
In this work, we propose a lightweight learnable codebook module capable of storing and retrieving question-specific reflections, thereby enhancing LLM performance across diverse tasks.
\section{Method}
In this section, we first present the process of LLM-based reflection generation in Section~\ref{method:ref_gen}. Next, we describe our proposed implicit feedback-free reflection approach in Section~\ref{method:feedbackfree_ref}. Subsequently, we introduce the concept of adaptive meta-reflection alignment in Section~\ref{method:align}. Finally, the overall optimization stage and inference stage are outlined in Section~\ref{method:train_infer}. The pipeline of \our is illustrated in Figure~\ref{fig:method}.

\subsection{LLM-based Reflection Generation}\label{method:ref_gen}
\begin{figure*}
    \centering
    \includegraphics[width=1\linewidth]{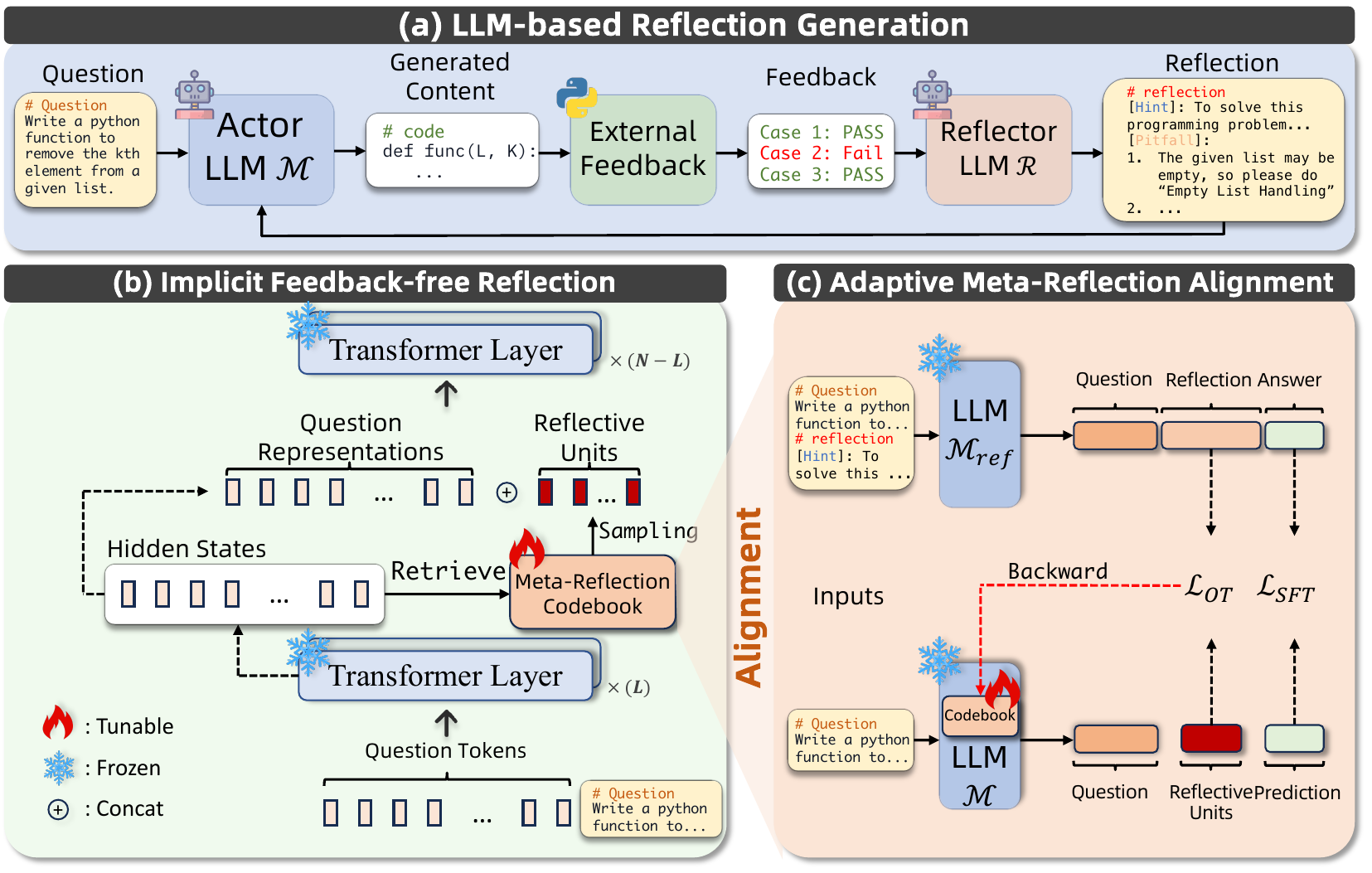}
    \caption{Overview of the Meta-Reflection framework: \textbf{(a)} LLM-based reflection generation through iterative processes; \textbf{(b) }Implicit feedback-free reflection, storing and retrieving reflective insights in a codebook; \textbf{(c)} Adaptive Meta-Reflection Alignment, incorporating reflective insights into the codebook.}
    \label{fig:method}
    \vspace{-0.8em}
\end{figure*}

Formally, consider a dataset \(U=\{(x,y)\}^{N}_{i=1}\), where~\(x\) represents a question and~\(y\) represents its corresponding answer. An actor LLM agent~\(\mathcal{M}\) is used to generate an initial output~\(\hat{y}_{\text{act}} =\mathcal{M}{(x)}\). However, this process may lead to unfaithful reasoning or hallucination~\cite{pan2023automatically}. To address these issues, reflection methods~\cite{,shinn2023reflexion,madaan2024self} propose leveraging feedback from external environment or golden labels~\cite{huang2024large} to refine the initial output~\(\hat{y}_{\text{act}}\). This feedback, denoted as~\(e = \mathcal{E}(x, \hat{y}_{\text{act}})\) where~\(\mathcal{E}\) represents the environment, provides comprehensive assessment of the initial output. For instance, in programming tasks, feedback typically includes interpreter information or execution results, while for mathematical problems, it involves comparing outputs against correct answer~\(y\).
Based on the feedback \(e\), a reflector LLM agent \(\mathcal{R}\) generates reflections \(r= \mathcal{R}(x,e)\), which guide the actor model \(\mathcal{M}\) to produce refined responses \(\hat{y}_{\text{ref}}=\mathcal{M}(x, r)\). 
As shown in Figure~\ref{fig:method}(a), this iterative process of generation, reflection, and refinement aims to enhance the quality and accuracy of the actor model~\(\mathcal{M}\)'s outputs, mitigating potential errors and improving overall performance~\cite{pan2023automatically}. 
Throughout the reflection generation process, we systematically curate a new dataset~\({D}_{t}=\{(x,r,\hat{y}_{\text{ref}})\}^{N^{\prime}}_{i=1}\) containing reflection-question-answer triplets. Details and corresponding prompts are provided in the Appendix.

\subsection{Implicit Feedback-free Reflection}\label{method:feedbackfree_ref}
As discussed in Section \ref{sec:intro}, reflection methodologies, while promising, are limited by their reliance on external feedback \cite{huang2024large} and computationally intensive multi-agent inference processes, hindering practical deployment.
Inspired by the adage "One never falls into the same ditch twice," which suggests that people learn from past mistakes without repeated feedback, we propose implicit feedback-free reflection. As shown in Figure \ref{fig:method}(b), this approach uses a learnable meta-reflection codebook to store and retrieve reflective insights, enabling efficient, feedback-free inference.

\paragraph{Meta-Reflection Codebook.}
The meta-reflection codebook consists of implicit reflective units~\({\boldsymbol{P}}\in \mathbb{R}^{K\times C}\),  where~\(K\) and~\(C\) denote codebook length and feature dimension, respectively. The question~\(x\) serves as the query to retrieve the relevant reflective units from the codebook.
Previous studies have demonstrated that intermediate layer features can provide sufficient preliminary understanding of input samples~\cite{2020deebert,zhang2024hyperllava}. Leveraging this insight, we utilize query representations from intermediate LLM layers, which contain rich semantic information for effective retrieval. Specifically, we position the meta-reflection codebook at the \(L\)-th layer~(\(0<L<N\)), where~\(N\) is the total layers of LLM, serving as a repository of reflective insights. To retrieve relevant reflective insights, the query is processed through the initial~\(L\) layers, transforming it into hidden states~\(\boldsymbol{H}^{L}_\text{query}\). We subsequently employ mean pooling~\(\mathcal{P}_\text{mean}\) to derive sentence-level representation as follows:
\begin{equation}
    \boldsymbol{h}=\mathcal{P}_\text{mean}(\boldsymbol{H}^{L}_\text{query}) \in \mathbb{R}^{1 \times C}
\end{equation}
The representation of the query is utilized to compute relevance score through:
\begin{equation}
    \mathbf{s}=\sigma(\frac{g(\boldsymbol{h}) f(\boldsymbol{P}^{T})}{\sqrt{K}})\in\mathbb{R}^{\text{1} \times K},
\end{equation}
where~\(\sigma\) denotes the softmax function, and \(g(\cdot)\) and~\(f(\cdot)\) represent transformation functions implemented as two-layer MLPs, which serve to stabilize the training process~\cite{liu2022ptuning}. The resulting score~\(\mathbf{s}\) quantifies the relevance between the question and reflective units from codebook, with higher scores indicating more applicable reflective units for the given query. Based on the score~\(\mathbf{s}\), we select the top-\(k\) relevant reflection units from the codebook to form the sequence~\(\boldsymbol{\hat{P}}_\text{ref}\in \mathbb{R}^{k \times C}\), maintaining their relative positions in the codebook. The concatenated sequence~\(\{\boldsymbol{H}^{L}_\text{query};\hat{\boldsymbol{P}}_\text{ref}\}\) is fed into the remaining~\((N-L)\) layers, incorporating question-specific reflective insights that guide the LLM's solution approach and enhance its performance. Notably, during the training phase, only the meta-reflection codebook is tunable while the backbone model remains frozen.

\paragraph{Sampling Strategy.}

To address the non-differentiable top-$k$ function that impedes gradient back-propagation during training, and to enhance the sampling diversity, we employ Gumbel-Softmax technique~\cite{jang2017categorical} with additional tricks~\cite{bengio2013estimating} to derive the sampling process:
\begin{gather}
    \hat{\mathbf{s}}=\sigma
    (\log(\mathbf{s})+\epsilon_{\text{gumbel}}) \in \mathbb{R}^{1\times K}, \nonumber \\ \label{equ:sampling}
    I=\mathds{1}_{i\in topk(\hat{s})}-sg[\hat{\mathbf{s}}]+\hat{\mathbf{s}} \in \mathbb{R}^{1\times K},
\end{gather}
where~\(\epsilon_{\text{gumbel}}  \in  \mathbb{R}^{1\times K}\) represents the Gumbel noise,~\(sg[\cdot]\) denotes the stop gradient operator and~\(\mathds{1}_{i\in topk(\hat{s})}\) indicates whether an index belongs to the top-\(k\) indices. The resulting indicator vector~\(I\) identifies the selected reflective units. This strategy ensures both differentiability during training and diverse sampling of reflective units.

\subsection{Adaptive Meta-Reflection Alignment}\label{method:align}

After acquiring the dataset \({D}_{t}\) as outlined in Section \ref{method:ref_gen}, our objective is to effectively leverage the information encapsulated within reflection \(r\). As depicted in Figure~\ref{fig:method}(c), we employ a same frozen LLM but with different input as the teacher model~\(\mathcal{M}_{\text{ref}}\), to process the input sequence \(\{x,r\}\) and extract the hidden states for each layer, \(\{\boldsymbol{P}^{l}_{\text{que}},\boldsymbol{P}^{l}_\text{ref}\}^{N}_{l=1}\), where \(\boldsymbol{P}^{l}_\text{que}\) and~\(\boldsymbol{P}^{l}_\text{ref}\) denote the hidden states of query and reflection sequences, respectively. The reflective units selected from the codebook are integrated into the final \(N-L\) layers, yielding \(\{\hat{\boldsymbol{P}}^{l}_\text{ref} \}^{N}_{l=L}\), with the purpose of aligning \(\{\boldsymbol{P}^{l}_\text{ref}\}^{N}_{l=L}\) and thereby embedding valuable information into the meta-reflection codebook. 
However, the dimensional variations and semantic misalignment between the ground-truth reflection \(\boldsymbol{P}^{l}_\text{ref}\) and the reflective units \(\hat{\boldsymbol{P}}^{l}_\text{ref}\) pose challenges for precise alignment between these sequences. To overcome this, we employ the optimal transport (OT) algorithm~\cite{rubner2000earth,liu2020self,Zhang_2020_CVPR}, which applies the earth mover’s distance (EMD) to gauge the semantic discrepancy between these two sequences.

\paragraph{OT for Meta-Reflection Alignment.} 
The EMD quantifies the distance between two discrete distributions as the minimum cost of transporting piles of dirt from "suppliers" to "demanders"~\cite{rosa}, framed as a linear optimization problem. Specifically, at the~\(l\)-th \((L<l<N)\) layer, we measure the distance required to transform \(\hat{\boldsymbol{P}}^{l}_\text{ref}\in \mathbb{R}^{k^{\prime}\times C}\) to \(\boldsymbol{P}^{l}_\text{ref}\in \mathbb{R}^{k\times C}\). Let each unit~\(\hat{\boldsymbol{p}}_{i} \in \hat{\boldsymbol{P}}^{l}_\text{ref}\) possesses a total of~\(\boldsymbol{r}_i\) quantities to transport, and each unit~\(\boldsymbol{p}_j \in \boldsymbol{P}^{l}_\text{ref}\) requires~\(\boldsymbol{c}_j\) quantities, forming the transport prototype:
\begin{gather}
\Pi(\boldsymbol{r},\boldsymbol{c}) = \{
\mathbf{\Gamma} \in \mathbb{R}^{k^{\prime} \times k}|\mathbf{\Gamma} \mathbf{1}_{k} = \boldsymbol{r},\boldsymbol{\Gamma}^T \mathbf{1}_{k^{\prime}}=\boldsymbol{c}\},
\end{gather}
where \(\boldsymbol{r}\in \mathbb{R}^{k^{\prime}}\) and~\(
\boldsymbol{c} \in \mathbb{R}^{k}\) are marginal weights for transportation matrix \(\mathbf{\Gamma}\) respectively. \(\mathbf{1}\) is all-one vector with corresponding size, and~\(\Pi(\boldsymbol{r},\boldsymbol{c})\) is the set of all possible distributions whose marginal weights are~\(\boldsymbol{r}\) and \(\boldsymbol{c}\).

We define the cost per unit transported from supplier token~\(\hat{p}_i\) to demander token~\(p_j\) as:
\begin{equation}
    \mathbf{D}_{ij} = 1 - \frac{{\boldsymbol{\hat{p}}^T_i}{\boldsymbol{p}_j}}{||\boldsymbol{\hat{p}}_i||||\boldsymbol{p}_j||},
\end{equation}
where tokens with similar representations incur lower transport costs. Given this, we can define the linear optimization problem as follows:
\begin{gather}~\label{equ:ot}
     \mathcal{R_{\text{OT}}}(\boldsymbol{r},\boldsymbol{c})= \min\limits_{\boldsymbol{\Gamma \in \Pi(r,c)}}   \sum\limits^{k^{\prime}}_{i}\sum\limits^{k}_{j}\mathbf{D}_{ij}\mathbf{\Gamma}_{ij}
\end{gather}

However, The exact minimization over~\(\mathbf{\Gamma}\) is solved in polynomial time and can be computationally intractable~\cite{arjovsky2017wasserstein,genevay2018learning}. Therefore, to find the optimal~\(\tilde{\mathbf{\Gamma}}\), we utilize \emph{Sinkhorn Algorithm}~\cite{Cuturi_2013} as an efficient approximation method. The detailed algorithm and the optimization process are shown in Appendix~\ref{app:appro_algo}. With optimal transportation matrix~\( \tilde{\mathbf{\Gamma}}\), the corresponding alignment loss for layer~\(l\) is:
\begin{equation}~\label{equ:loss}
    \mathcal{L}^{l}_\text{OT}=\langle \tilde{\mathbf{\Gamma}}, \mathbf{D}\rangle_\mathbf{\mathrm{F}},
\end{equation}
and the overall alignment loss is calculated as the mean across the last~\(N-L\) layers:
\begin{equation}
    \mathcal{L}_\text{OT}=\frac{\sum^{N}_{l=L}{\mathcal{L}^{l}_\text{OT}}}{N-L}\ \label{equ:all_layers_loss}
\end{equation}

The alignment loss quantifies the semantic gap~\cite{li2020improving} between the reflective units from the meta-reflection codebook and the ground-truth reflection. In our scenario, by minimizing the~\(\mathcal{L}_\text{OT}\), the reflective insights from ground-truth reflection are incorporated into the codebook, enhancing the model~\(\mathcal{M}\)'s capacity to handle complex tasks and improve overall performance.

\subsection{Optimization and Inference} \label{method:train_infer}
We delineate the overall optimization and inference stages as follows:

\paragraph{Progressive Optimization Stage.}
We employ a progressive optimization paradigm to enhance model performance. Initially, we utilize \(\mathcal{L}_\text{OT}\) to align the reflective units from codebook with ground truth reflections, infusing reflective information into the codebook of the model~\(\mathcal{M}\). Subsequently, we leverage labels from dataset \(\mathcal{D}_t\) to fine-tune the codebook using the vanilla supervised learning loss \(\mathcal{L}_\text{SFT}\). This optimization paradigm ensures stable training progression and effective incorporation of reflective information, enhancing the model's ability to capture and utilize this knowledge while maintaining overall learning stability.

\paragraph{Inference Stage.}
During the inference stage, the input question~\(x\) serves as query to retrieve pertinent reflective units from the meta-reflection codebook, guiding the LLM in addressing complex tasks. The retrieval process, elucidated in Section~\ref{method:feedbackfree_ref}, is executed only once at the generation of the initial token. Leveraging the characteristics of causal language models, this inference stage can also utilize KV caching~\cite{pope2023efficiently} to mitigate computational overhead. 

\section{Experiments}
\begin{table*}[!htb]
  \centering
  \scalebox{0.8}{
      \begin{tabular}{lccccccccc}
          \toprule
          & &\multicolumn{4}{c}{\textbf{MBPP}} & \multicolumn{4}{c}{\textbf{HumanEval}} \\
          \cmidrule(lr){3-6} \cmidrule(lr){7-10}
          & & \multicolumn{2}{c}{\textbf{LLaMA-3.1}} &\multicolumn{2}{c}{\textbf{CodeLlama}} &\multicolumn{2}{c}{\textbf{LLaMA-3.1}} &\multicolumn{2}{c}{\textbf{CodeLlama}}\\
          \cmidrule(lr){3-4} \cmidrule(lr){5-6}\cmidrule(lr){7-8} \cmidrule(lr){9-10}
          \textbf{Methods}& \emph{ref} & \textbf{Pass @ 1} & \textbf{Pass @ 3} & \textbf{Pass @ 1} & \textbf{Pass @ 3} & \textbf{Pass @ 1} & \textbf{Pass @ 3} & \textbf{Pass @ 1} &\textbf{Pass @ 3}\\
          \midrule
          $\text{Zero-Shot}$&\xmark & 58.8&  68.0& 40.4&  49.2&62.7& 68.3&41.0&47.8\\
          $\text{Few-Shot}$&\xmark &59.6& 68.6&41.4& 50.6& 63.4& 70.8&42.2&48.5\\
          \midrule
          $\text{LoRA}$& \xmark&60.4&  69.0& 41.6&  54.2& 62.1& 72.1&\underline{43.5}&52.8\\
          $\text{P-Tuning}$ &\xmark&59.4&  68.8&42.8&  55.6& 62.1& 73.3&42.9&52.2\\
          $\text{Llama-Adapter}$& \xmark & 59.6& 68.2& \underline{45.4}& \underline{56.0}& 62.7& 73.3& 42.9&53.4\\
          \midrule
          $\text{Re-ReST}$&\cmark& \underline{60.2}&  \underline{69.6}& 42.4&  55.2&\underline{63.4}& \underline{73.9}&42.2&\underline{53.4}\\
          $\text{Reflection}_{(\text{RAG})}$&\cmark & 58.6&  67.2&41.2&  51.2& 62.7& 67.1&35.4&46.6\\
          \midrule
          \rowcolor{Gray} \text{Ours}&\cmark & \textbf{63.4}& \textbf{70.4}& \textbf{46.8}& \textbf{57.6}&\textbf{64.6}& \textbf{75.2}&\textbf{45.3}&\textbf{55.9}\\
          \bottomrule
      
      \end{tabular}
  }
  \caption{The experimental results on two programming benchmarks: MBPP~\cite{austin2021program} and HumanEval~\cite{chen2021evaluating} datasets. We report the performance using Pass@1 and Pass@3 metrics. Here, \emph{ref} indicates the utilization of reflection mechanism. The \textbf{boldface} and \underline{underline} fonts denote the best and second-best performance, respectively.}
  \label{tab:programming}
\end{table*}

\begin{table*}[!htb]
    \begin{minipage}[t]{0.47\textwidth}  
        \centering
        \scalebox{0.85}{
        \begin{tabular}{l|ccc}
          \toprule
          \textbf{Methods} & \emph{ref} & \textbf{LLaMA-3.1}& \textbf{Qwen-2} \\
          \midrule
          $\text{Zero-Shot}$&\xmark& 78.4 & 78.1\\
          $\text{Few-Shot}$&\xmark & 80.4 &79.5\\
          \midrule
          $\text{LoRA}$ &\xmark & 80.7 & 80.0\\
          $\text{P-Tuning}$ &\xmark & 79.4 & 79.6\\
          \midrule
          $\text{Re-ReST}$ & \cmark& \underline{82.4}&\underline{84.8}\\
          $\text{Reflection}_{(\text{RAG})}$& \cmark& 77.7 &76.7\\
          \midrule
          \rowcolor{Gray} \text{Ours}&\cmark &\textbf{85.3}& \textbf{86.7}\\
          \bottomrule
        \end{tabular}
        }
        \caption{The experimental results on a mathematical reasoning benchmark: GSM8K~\cite{cobbe2021gsm8k}.}
        \label{tab:math}
    \end{minipage}
    \hfill
    \begin{minipage}[t]{0.47\textwidth}
        \centering
        \scalebox{0.85}{
        \begin{tabular}{l|ccc}
          \toprule
          \textbf{Methods} & \emph{ref} & \textbf{LLaMA-3.1} & \textbf{Qwen-2} \\
          \midrule
          $\text{Zero-Shot}$&\xmark& 83.5& 89.8\\
          $\text{Few-Shot}$&\xmark & 85.5& 90.8\\
          \midrule
          $\text{LoRA}$ &\xmark & \underline{86.9}& \underline{91.1}\\
          $\text{P-Tuning}$ &\xmark & 85.5& 90.9\\
          \midrule
          $\text{Re-ReST}$ & \cmark& 85.5& 90.9\\
          $\text{Reflection}_{(\text{RAG})}$& \cmark& 81.8& 86.6\\
          \midrule
          \rowcolor{Gray} \text{Ours}&\cmark &\textbf{89.7}& \textbf{92.9}\\
          \bottomrule
        \end{tabular}
        }
        \caption{The experimental results on ECID dataset in E-commerce domain.}
        \label{tab:ecid}
    \end{minipage}
    \vspace{-1em}

\end{table*}

\subsection{Datasets}\label{exp:dataset}
We assess our method on diverse datasets across different domains: programming (\ie, MBPP, HumanEval), mathematical reasoning (\ie, GSM8K), and E-commerce customer intent detection (\ie, ECID). Details can be found in Appendix~\ref{app:ecid} and \ref{app:datasets}.

\paragraph{Programming.}
We evaluate our approach on two Python code generation benchmarks (MBPP~\cite{austin2021program} and HumanEval~\cite{chen2021evaluating}), using Pass@k metric to measure the percentage of problems that successfully pass all unit tests within k attempts~\cite{dou2024reflection}.

\paragraph{Mathematical Reasoning.}
For mathematical reasoning task, We employ the Grade School Math 8K (GSM8K) dataset~\cite{cobbe2021gsm8k} for evaluating~\our. We utilize the Exact Match (EM) metric between the generated response and the correct answer~\cite{madaan2024self}.

\paragraph{E-commerce Customer Intent Detection (ECID).}
Intelligent Customer Service (ICS) in e-commerce has emerged as a prominent application for Large Language Models (LLMs)~\cite{kolasani2023optimizing}. Research by~\citet{cheng2024impact} has highlighted that the key to enhancing ICS performance lies in accurately inferring customers' core service intent through the analysis of historical customer-agent interactions and corresponding order data. However, current LLMs struggle with precise intent detection, primarily due to semantic ambiguities inherent in diverse service requests.
To evaluate the efficacy of our proposed approach in this domain, we introduce the E-commerce Customer Intent Detection (ECID) dataset. This dataset comprises meticulously cleaned and systematically labeled Chinese language data from Taobao online customer service interactions, resulting in 1,170 high-quality entries. Details of the ECID can be found in Appendix~\ref{app:ecid}.

\begin{figure*}[htpb]  
    \centering
    \begin{minipage}{0.32\textwidth}
        \centering
        \small
              \begin{center}
                \includegraphics[width=\linewidth]{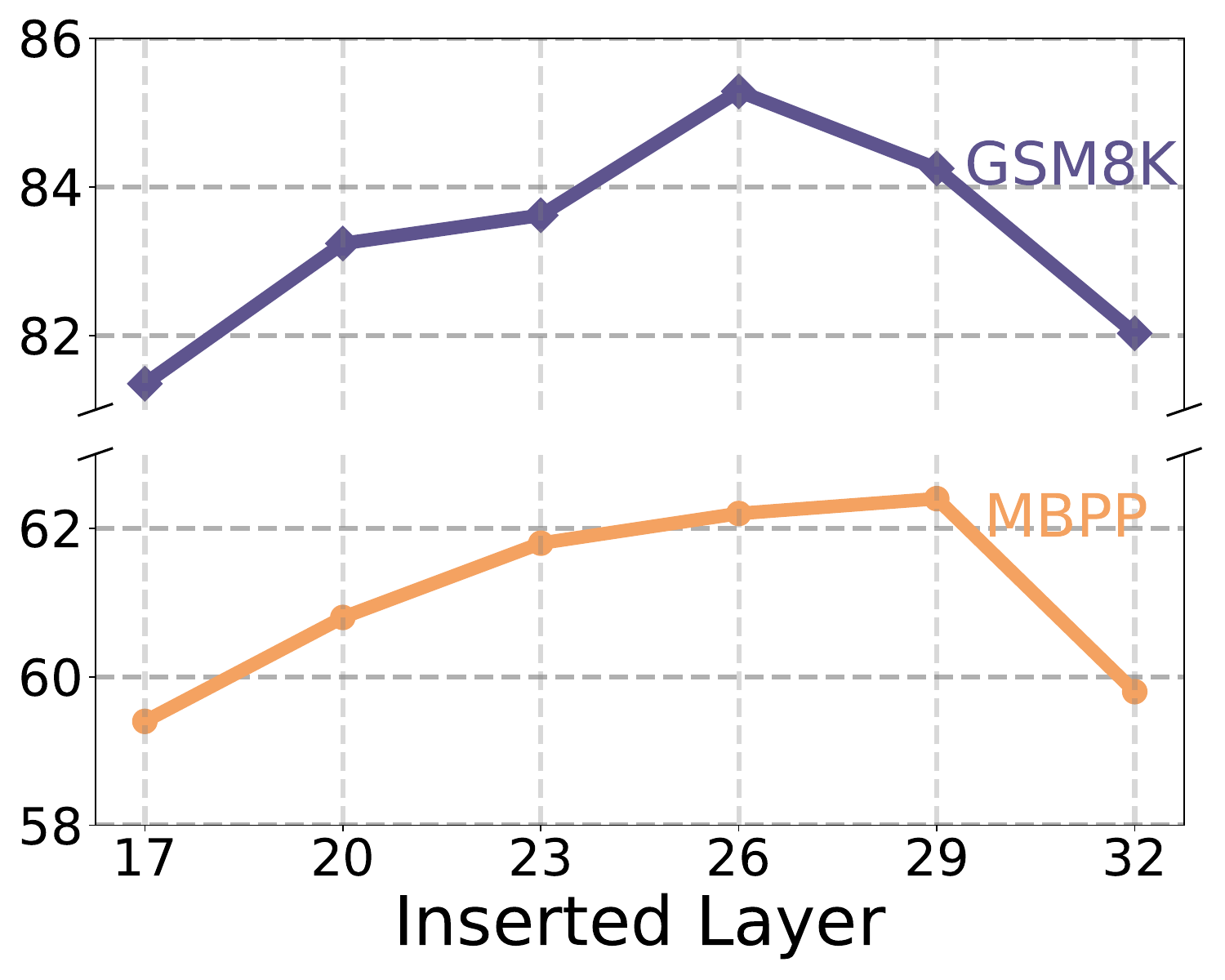}
              \end{center}
    \end{minipage}
    \hfill
    \begin{minipage}{0.32\textwidth}
        \centering
        \small
              \begin{center}
                \includegraphics[width=\linewidth]{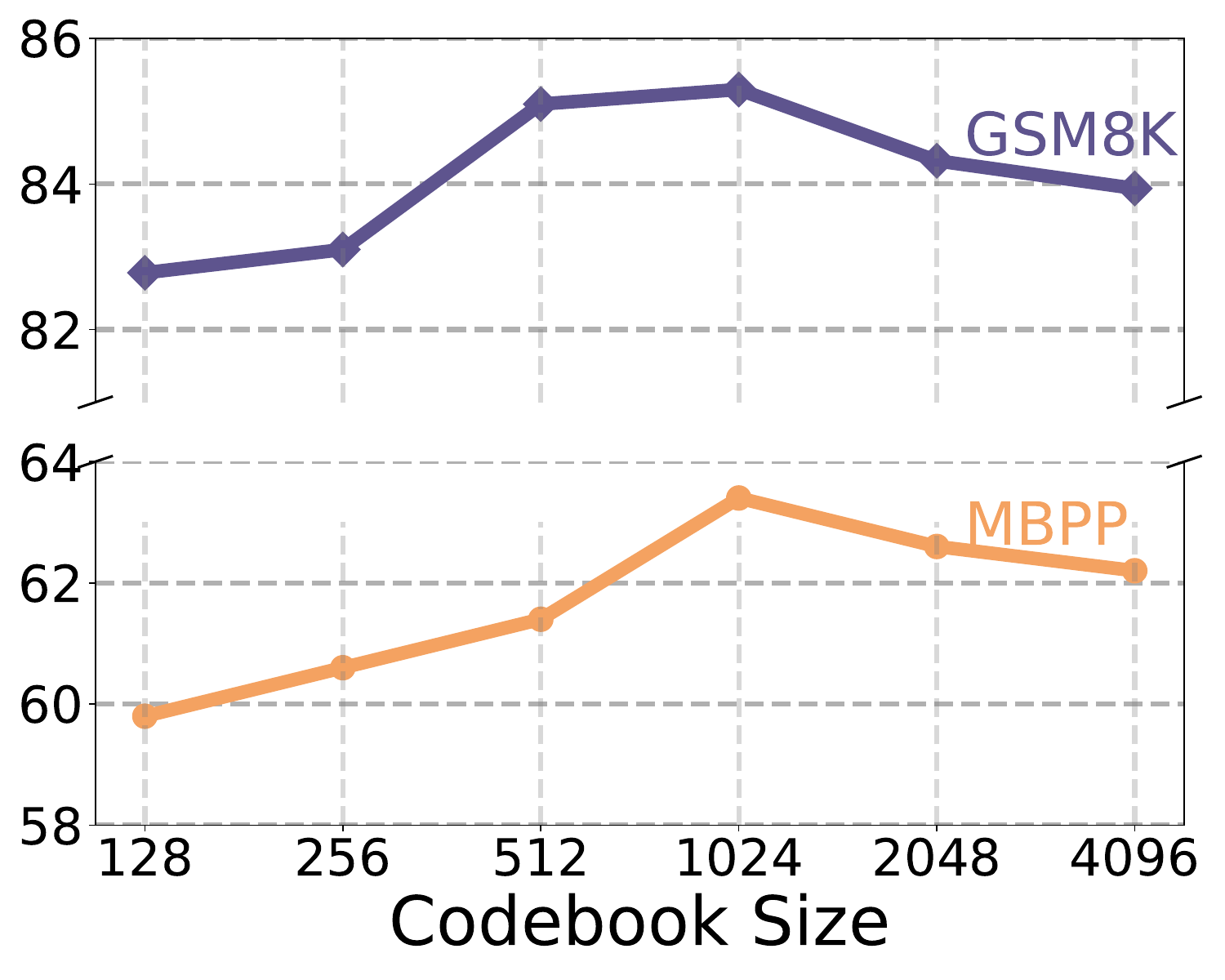}
              \end{center}
    \end{minipage}
    \hfill
    \begin{minipage}{0.32\textwidth}
        \centering
        \small
              \begin{center}
                \includegraphics[width=\linewidth]{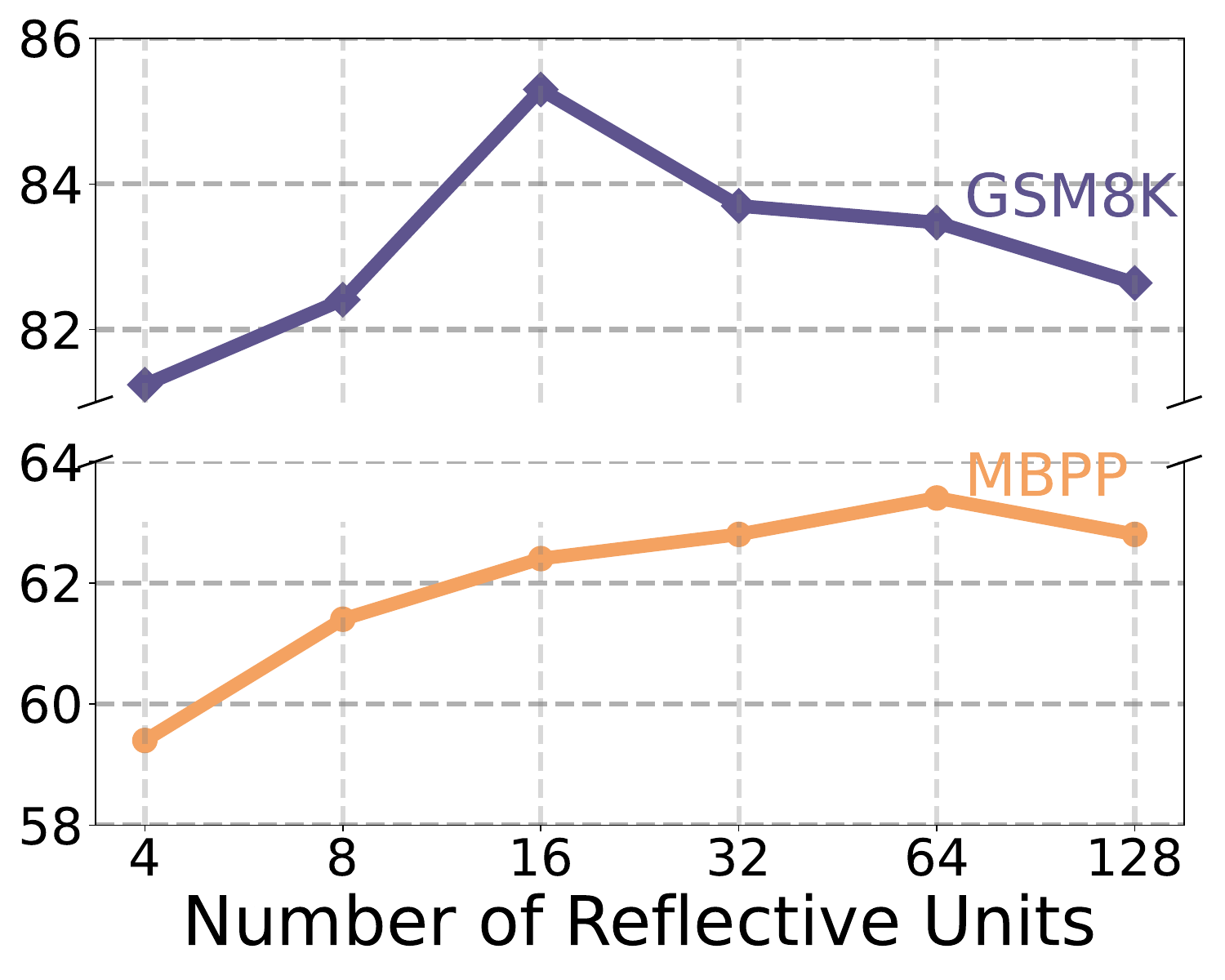}
              \end{center}
    \end{minipage}
    \caption{
  Sensitivity analysis of three critical hyper-parameters: \textbf{Left}: Insertion layer position of the codebook; \textbf{Middle}: Total number of reflective units in codebook; \textbf{Right}: Number of reflective units selected per inference.}
    \label{fig:sensitive}
    \vspace{-0.5em}
\end{figure*}

\subsection{Experimental Setup}\label{exp:setup}
\paragraph{Models.} We evaluate~\our across various open-source LLMs. For the actor models, we utilize \texttt{Qwen-2-7B-Instruct}~\cite{yang2024qwen2}, \texttt{Llama-3.1-8B-Instruct}~\cite{dubey2024llama}, and \texttt{CodeLlama-7B-Instruct}~\cite{roziere2023code}. \texttt{Qwen-2-72B-Instruct} serves as the reflector model.

\paragraph{Baselines.}
To evaluate the effectiveness of our proposed method, we compare it with three types of baselines:
\textbf{Common reasoning}: including Zero-Shot and Few-Shot approaches.
\textbf{PEFT methods}: Adapter-based approaches such as LoRA~\cite{hu2021lora} and Llama-Adapter~\cite{zhang2023llama}, as well as Prompt-based methods like P-Tuning~\cite{liu2022ptuning}.
\textbf{Reflection-based methods}: Re-ReST~\cite{dou2024reflection} for reflection-enhanced training. Additionally, we implement Reflection-RAG, which generates reflections on training data and employs Retrieval-Augmented Generation (RAG)~\cite{gao2023retrieval} during inference to select the most relevant question-specific reflections. The details of baselines are in the Appendix~\ref{app:baseline}.

\subsection{Main Results}\label{exp:result}
Tables~\ref{tab:programming}, \ref{tab:math}, and \ref{tab:ecid} present the experimental results across three distinct domains: programming, mathematical reasoning, and e-commerce customer intent detection. 

Our empirical investigation reveals fundamental limitations in base LLMs' domain-specific capabilities, as demonstrated by CodeLlama's modest 40.4\% performance on MBPP under the Pass@1 metric. This deficiency primarily stems from these models' \textbf{insufficient domain knowledge and capabilities}. While tuning with PEFT methods like LoRA demonstrate potential for improvement, the gains remain incremental—yielding mere 1.2\% and 0.2\% improvements in Zero-Shot and Few-Shot settings respectively. This suggests that current supervised learning paradigms, while domain knowledge internalization during finetuning, \textbf{fail to address the critical need for guidance during inference.}

Recent advances in reflection-based methodologies, particularly Re-ReST, have shown promise by implicitly incorporating reflective guidance through refined self-training data, evidenced by LLaMA-3.1's 1.7\% performance improvement over LoRA on GSM8K. However, these approaches still \textbf{neglect the crucial aspect of explicit, granular guidance during the inference phase}. Although leveraging RAG-retrieved reflections as explicit guidance appears promising, empirical results on benchmarks like GSM8K and ECID demonstrate suboptimal performance even compared to common reasoning approaches. This degradation occurs because retrieved reflections, though relevant to source problems, often \textbf{lack precise applicability to similar cases and may introduce noise}, particularly in mathematical tasks requiring fine-grained guidance. Comprehensive case studies supporting these findings are presented in Appendix~\ref{app:case_study}. Our proposed methodology addresses these limitations by providing explicit, fine-grained reflective guidance during inference, significantly outperforming existing approaches across all baseline metrics.

\subsection{Inference Efficiency Analysis}\label{exp:eff}
\begin{table}[t]
    \centering
    \scalebox{0.83}{
    \begin{tabular}{lccc}
    \toprule
     \multirow{2}{*}{\textbf{Methods}}& \multicolumn{3}{c}{\textbf{First Token Latency} ($\downarrow$)} \\
     \cmidrule(lr){2-4}
     &\text{Retrieve} &\text{LLM Processing}& \text{Total} \\
     \midrule
    \text{Zero-Shot} &$-$ & 149 ms& 149 ms\\
 \text{Few-Shot}& $-$& 545 ms& 545 ms\\
    $\text{Reflection}_{(\text{RAG})}$ & 642 ms&309 ms& 951 ms\\
     \midrule    
     \rowcolor{Gray} Ours & 93 ms&153 ms&246 ms\\
     \bottomrule
    \end{tabular}}
    \caption{We analyze inference efficiency on the \texttt{ECID} dataset by measuring the first token (in milliseconds). The first token latency is decomposed into retrieval time and LLM processing time. All measurements are conducted using a 24-core Intel(R) Xeon(R) Platinum 8163 CPU @ 2.50GHz and 2 NVIDIA Tesla V100 GPUs.}
    \label{tab:efficiency}
    \vspace{-1.3em}
\end{table}

We evaluate the inference efficiency of \our, with results presented in Table~\ref{tab:efficiency}. Compared to existing reflection-based methods like Reflection-RAG that require separate encoders and knowledge base retrieval, our approach leverages LLM's intermediate layer representations for retrieval.
Furthermore, RAG-based methods store knowledge in a discrete format, necessitating a large-scale knowledge base. In contrast, \our captures knowledge and reflective insights in a dense format, enabling the construction of a smaller, more compact knowledge base, thereby significantly reducing computational overhead.
Notably, our method achieves comparable first-token latency to common reasoning approaches while maintaining the benefits of reflection-based reasoning, demonstrating its practicality for real-world applications.

\subsection{Sensitive Analysis}\label{exp:sen}
We perform sensitivity analysis on three critical hyper-parameters of~\our: inserted layers, codebook size, and number of reflective units. The Experimental results are presented in Figure~\ref{fig:sensitive}.

\paragraph{Inserted Layer.}
The positioning of the meta-reflection codebook layer critically influences the balance between retrieval quality and reflective information integration. Analysis from Figure~\ref{fig:sensitive} (Left) reveals that early-layer insertion results in insufficient semantic query encoding, while late-layer placement constrains the processing of retrieved reflective components. Our empirical results demonstrate that an intermediate-posterior position (\eg, layer 26) achieves optimal performance.

\paragraph{Codebook Size.}
The codebook size, which represents the total number of reflective units, determines the capacity of the codebook.
As shown in Figure~\ref{fig:sensitive} (Middle), we observe that a codebook size of 1024 yields optimal performance. Smaller sizes may lead to underfitting, while larger sizes can result in a sparse codebook, potentially causing training instability. 
 
\paragraph{Number of Reflective Units.}
As illustrated in Figure~\ref{fig:sensitive} (Right), the optimal number of reflective units varies proportionally with task complexity. Notably, while base LLaMA-3.1 achieves a substantial 78.4\% performance on GSM8K, the inherently more challenging MBPP dataset requires additional reflective insights to provide comprehensive guidance. This observation underscores the relationship between task complexity and the requisite quantity of reflective support.

\begin{figure}[t]
    \centering
    \includegraphics[scale=0.4]{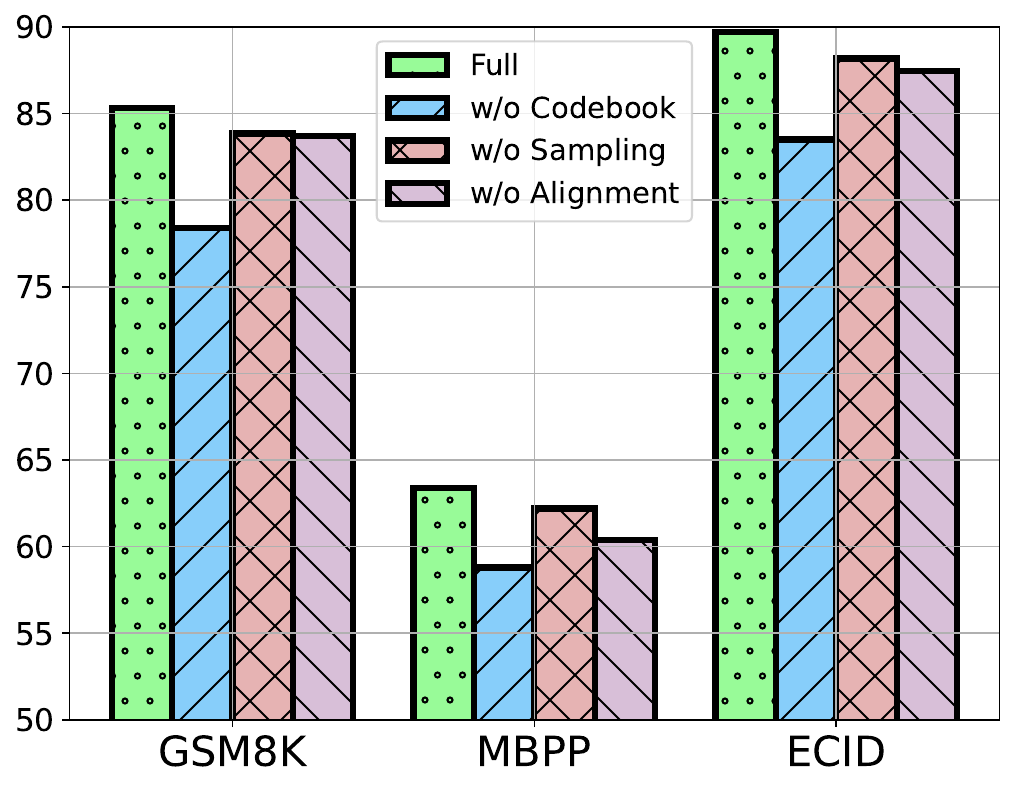} 
    \vspace{-0.5em}
    \caption{Experimental results of ablation study.}
    \label{fig:ablation}
    \vspace{-1.2em}
\end{figure}

\subsection{Ablation Study}\label{exp:ablation}

In this section, we conduct a comprehensive ablation study to evaluate the impact of various components in~\our. We examine three key variants: `w/o Codebook' (no meta-reflection codebook), `w/o Sampling' (no sampling strategy defined in Equation~\ref{equ:sampling}), and `w/o Alignment' (no alignment mechanism described in Equation~\ref{equ:all_layers_loss}).
As illustrated in Figure~\ref{fig:ablation}, the meta-reflection codebook demonstrates significant effectiveness in storing and retrieving reflective units that guide LLMs through the problem-solving process. The ablation analysis further demonstrates that both the sampling strategy and meta-reflection alignment mechanism play crucial roles in maintaining solution diversity and incorporating reflective insights, respectively, thereby enhancing overall performance.

\subsection{Visualization}\label{exp:vis}

\begin{figure}[t]
    \centering
    \includegraphics[width=1\linewidth]{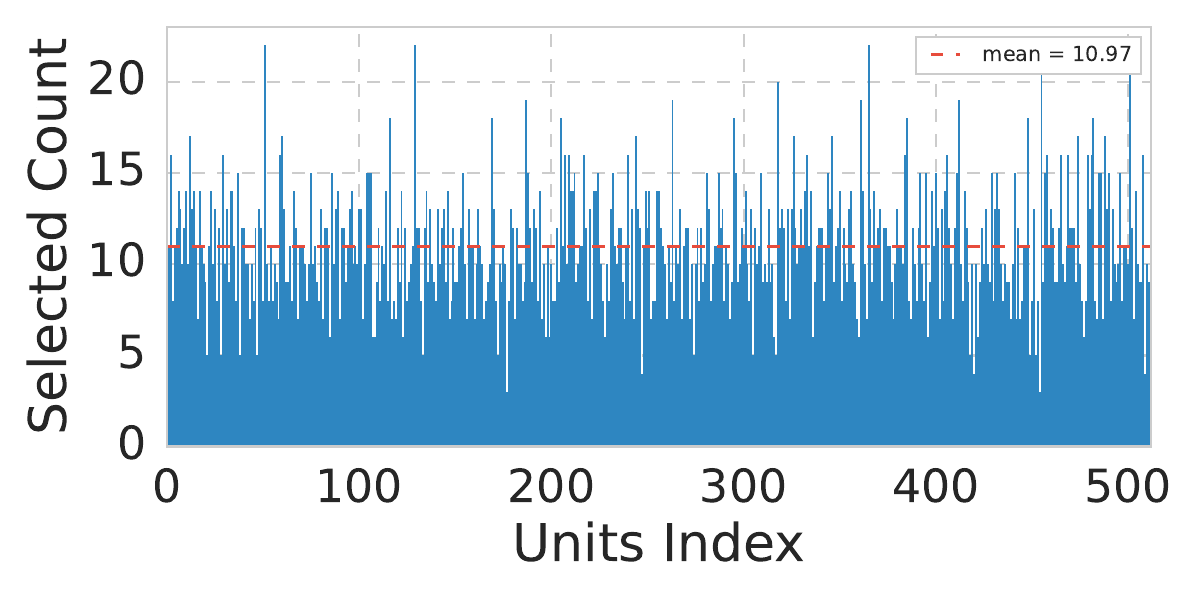}
    \vspace{-2em}
    \caption{Visualization of reflective unit selection frequencies distribution in the ECID dataset. The \textbf{x-axis} represents the unit indices, while the \textbf{y-axis} shows their cumulative selection counts during inference. The codebook is configured with a size of 512 units, with 16 units selected per inference.}
    \vspace{-1em}
    \label{fig:vis_unit_count_distribution_ecid}
\end{figure}

We visualize the selection frequency distribution of reflective units within the meta-reflection codebook. As shown in Figure~\ref{fig:vis_unit_count_distribution_ecid}, the selection patterns of reflective units vary significantly. Notably, certain units exhibit higher selection frequencies, potentially reflecting commonly applicable insights, whereas others are selected less frequently, suggesting their specialized nature. Additional visualization results are provided in Appendix~\ref{app:vis}.

\section{Conclusion}
In this paper, we introduce \our, a novel feedback-free reflection mechanism that operates with a single inference pass without requiring external feedback. Our approach incorporates reflective insights within a codebook structure, facilitating efficient storage, retrieval, and utilization of historical insights to guide LLMs in problem-solving tasks. 
To validate the practical applicability of our method, we propose a new industrial benchmark: E-commerce Customer Intent Detection (ECID). Comprehensive experiments conducted across diverse domains and the ECID benchmark demonstrate the effectiveness and efficiency of~\our.
\section{Limitations}
This work introduces ~\our, a novel feedback-free reflection mechanism that operates with a single inference pass without requiring external feedback. However, ~\our is primarily applicable to parameter-accessible LLMs (\eg, Qwen and LLaMA) and cannot be extended to models where parameters are inaccessible through API-only interfaces (\eg, ChatGPT and Claude).

\bibliography{custom, anthology}

\begin{thebibliography}{46}
\providecommand{\natexlab}[1]{#1}

\bibitem[{Achiam et~al.(2023)Achiam, Adler, Agarwal, Ahmad, Akkaya, Aleman, Almeida, Altenschmidt, Altman, Anadkat et~al.}]{achiam2023gpt}
Josh Achiam, Steven Adler, Sandhini Agarwal, Lama Ahmad, Ilge Akkaya, Florencia~Leoni Aleman, Diogo Almeida, Janko Altenschmidt, Sam Altman, Shyamal Anadkat, et~al. 2023.
\newblock Gpt-4 technical report.
\newblock \emph{arXiv preprint arXiv:2303.08774}.

\bibitem[{Arjovsky et~al.(2017)Arjovsky, Chintala, and Bottou}]{arjovsky2017wasserstein}
Martin Arjovsky, Soumith Chintala, and L{\'e}on Bottou. 2017.
\newblock Wasserstein generative adversarial networks.
\newblock In \emph{International conference on machine learning}, pages 214--223. PMLR.

\bibitem[{Austin et~al.(2021)Austin, Odena, Nye, Bosma, Michalewski, Dohan, Jiang, Cai, Terry, Le et~al.}]{austin2021program}
Jacob Austin, Augustus Odena, Maxwell Nye, Maarten Bosma, Henryk Michalewski, David Dohan, Ellen Jiang, Carrie Cai, Michael Terry, Quoc Le, et~al. 2021.
\newblock Program synthesis with large language models.
\newblock \emph{arXiv preprint arXiv:2108.07732}.

\bibitem[{Bengio et~al.(2013)Bengio, L{\'e}onard, and Courville}]{bengio2013estimating}
Yoshua Bengio, Nicholas L{\'e}onard, and Aaron Courville. 2013.
\newblock Estimating or propagating gradients through stochastic neurons for conditional computation.
\newblock \emph{arXiv preprint arXiv:1308.3432}.

\bibitem[{Brown et~al.(2020)Brown, Mann, Ryder, Subbiah, Kaplan, Dhariwal, Neelakantan, Shyam, Sastry, Askell, Agarwal, Herbert-Voss, Krueger, Henighan, Child, Ramesh, Ziegler, Wu, Winter, Hesse, Chen, Sigler, Litwin, Gray, Chess, Clark, Berner, McCandlish, Radford, Sutskever, and Amodei}]{fewshot}
Tom~B. Brown, Benjamin Mann, Nick Ryder, Melanie Subbiah, Jared Kaplan, Prafulla Dhariwal, Arvind Neelakantan, Pranav Shyam, Girish Sastry, Amanda Askell, Sandhini Agarwal, Ariel Herbert-Voss, Gretchen Krueger, Tom Henighan, Rewon Child, Aditya Ramesh, Daniel~M. Ziegler, Jeffrey Wu, Clemens Winter, Christopher Hesse, Mark Chen, Eric Sigler, Mateusz Litwin, Scott Gray, Benjamin Chess, Jack Clark, Christopher Berner, Sam McCandlish, Alec Radford, Ilya Sutskever, and Dario Amodei. 2020.
\newblock Language models are few-shot learners.
\newblock In \emph{Proceedings of the 34th International Conference on Neural Information Processing Systems}, NIPS '20, Red Hook, NY, USA. Curran Associates Inc.

\bibitem[{Chen et~al.(2021)Chen, Tworek, Jun, Yuan, de~Oliveira~Pinto, Kaplan, Edwards, Burda, Joseph, Brockman, Ray, Puri, Krueger, Petrov, Khlaaf, Sastry, Mishkin, Chan, Gray, Ryder, Pavlov, Power, Kaiser, Bavarian, Winter, Tillet, Such, Cummings, Plappert, Chantzis, Barnes, Herbert-Voss, Guss, Nichol, Paino, Tezak, Tang, Babuschkin, Balaji, Jain, Saunders, Hesse, Carr, Leike, Achiam, Misra, Morikawa, Radford, Knight, Brundage, Murati, Mayer, Welinder, McGrew, Amodei, McCandlish, Sutskever, and Zaremba}]{chen2021evaluating}
Mark Chen, Jerry Tworek, Heewoo Jun, Qiming Yuan, Henrique~Ponde de~Oliveira~Pinto, Jared Kaplan, Harri Edwards, Yuri Burda, Nicholas Joseph, Greg Brockman, Alex Ray, Raul Puri, Gretchen Krueger, Michael Petrov, Heidy Khlaaf, Girish Sastry, Pamela Mishkin, Brooke Chan, Scott Gray, Nick Ryder, Mikhail Pavlov, Alethea Power, Lukasz Kaiser, Mohammad Bavarian, Clemens Winter, Philippe Tillet, Felipe~Petroski Such, Dave Cummings, Matthias Plappert, Fotios Chantzis, Elizabeth Barnes, Ariel Herbert-Voss, William~Hebgen Guss, Alex Nichol, Alex Paino, Nikolas Tezak, Jie Tang, Igor Babuschkin, Suchir Balaji, Shantanu Jain, William Saunders, Christopher Hesse, Andrew~N. Carr, Jan Leike, Josh Achiam, Vedant Misra, Evan Morikawa, Alec Radford, Matthew Knight, Miles Brundage, Mira Murati, Katie Mayer, Peter Welinder, Bob McGrew, Dario Amodei, Sam McCandlish, Ilya Sutskever, and Wojciech Zaremba. 2021.
\newblock \href {https://arxiv.org/abs/2107.03374} {Evaluating large language models trained on code}.
\newblock \emph{Preprint}, arXiv:2107.03374.

\bibitem[{Cheng et~al.(2024)Cheng, Fan, Shao, Jia, and Zhang}]{cheng2024impact}
Zhendong Cheng, Wenfang Fan, Bingjia Shao, Wenli Jia, and Yong Zhang. 2024.
\newblock The impact of intelligent customer service agents’ initial response on consumers’ continuous interaction intention.
\newblock \emph{Journal of Retailing and Consumer Services}, 76:103585.

\bibitem[{Cobbe et~al.(2021)Cobbe, Kosaraju, Bavarian, Chen, Jun, Kaiser, Plappert, Tworek, Hilton, Nakano, Hesse, and Schulman}]{cobbe2021gsm8k}
Karl Cobbe, Vineet Kosaraju, Mohammad Bavarian, Mark Chen, Heewoo Jun, Lukasz Kaiser, Matthias Plappert, Jerry Tworek, Jacob Hilton, Reiichiro Nakano, Christopher Hesse, and John Schulman. 2021.
\newblock Training verifiers to solve math word problems.
\newblock \emph{arXiv preprint arXiv:2110.14168}.

\bibitem[{Cuturi(2013)}]{Cuturi_2013}
Marco Cuturi. 2013.
\newblock Sinkhorn distances: Lightspeed computation of optimal transport.
\newblock \emph{Neural Information Processing Systems,Neural Information Processing Systems}.

\bibitem[{Dou et~al.(2024)Dou, Yang, Wu, Chang, and Peng}]{dou2024reflection}
Zi-Yi Dou, Cheng-Fu Yang, Xueqing Wu, Kai-Wei Chang, and Nanyun Peng. 2024.
\newblock Reflection-reinforced self-training for language agents.
\newblock \emph{arXiv preprint arXiv:2406.01495}.

\bibitem[{Du et~al.(2023)Du, Li, Torralba, Tenenbaum, and Mordatch}]{du2023improving}
Yilun Du, Shuang Li, Antonio Torralba, Joshua~B Tenenbaum, and Igor Mordatch. 2023.
\newblock Improving factuality and reasoning in language models through multiagent debate.
\newblock \emph{arXiv preprint arXiv:2305.14325}.

\bibitem[{Dubey et~al.(2024)Dubey, Jauhri, Pandey, Kadian, Al-Dahle, Letman, Mathur, Schelten, Yang, Fan et~al.}]{dubey2024llama}
Abhimanyu Dubey, Abhinav Jauhri, Abhinav Pandey, Abhishek Kadian, Ahmad Al-Dahle, Aiesha Letman, Akhil Mathur, Alan Schelten, Amy Yang, Angela Fan, et~al. 2024.
\newblock The llama 3 herd of models.
\newblock \emph{arXiv preprint arXiv:2407.21783}.

\bibitem[{Gao et~al.(2023)Gao, Xiong, Gao, Jia, Pan, Bi, Dai, Sun, and Wang}]{gao2023retrieval}
Yunfan Gao, Yun Xiong, Xinyu Gao, Kangxiang Jia, Jinliu Pan, Yuxi Bi, Yi~Dai, Jiawei Sun, and Haofen Wang. 2023.
\newblock Retrieval-augmented generation for large language models: A survey.
\newblock \emph{arXiv preprint arXiv:2312.10997}.

\bibitem[{Genevay et~al.(2018)Genevay, Peyr{\'e}, and Cuturi}]{genevay2018learning}
Aude Genevay, Gabriel Peyr{\'e}, and Marco Cuturi. 2018.
\newblock Learning generative models with sinkhorn divergences.
\newblock In \emph{International Conference on Artificial Intelligence and Statistics}, pages 1608--1617. PMLR.

\bibitem[{Hu et~al.(2021)Hu, Wallis, Allen-Zhu, Li, Wang, Wang, Chen et~al.}]{hu2021lora}
Edward~J Hu, Phillip Wallis, Zeyuan Allen-Zhu, Yuanzhi Li, Shean Wang, Lu~Wang, Weizhu Chen, et~al. 2021.
\newblock Lora: Low-rank adaptation of large language models.
\newblock In \emph{International Conference on Learning Representations}.

\bibitem[{Hu et~al.(2023)Hu, Wang, Lan, Xu, Lim, Bing, Xu, Poria, and Lee}]{hu2023llm}
Zhiqiang Hu, Lei Wang, Yihuai Lan, Wanyu Xu, Ee-Peng Lim, Lidong Bing, Xing Xu, Soujanya Poria, and Roy Lee. 2023.
\newblock Llm-adapters: An adapter family for parameter-efficient fine-tuning of large language models.
\newblock In \emph{Proceedings of the 2023 Conference on Empirical Methods in Natural Language Processing}, pages 5254--5276.

\bibitem[{Huang et~al.(2024)Huang, Chen, Mishra, Zheng, Yu, Song, and Zhou}]{huang2024large}
Jie Huang, Xinyun Chen, Swaroop Mishra, Huaixiu~Steven Zheng, Adams~Wei Yu, Xinying Song, and Denny Zhou. 2024.
\newblock \href {https://openreview.net/forum?id=IkmD3fKBPQ} {Large language models cannot self-correct reasoning yet}.
\newblock In \emph{The Twelfth International Conference on Learning Representations}.

\bibitem[{Jang et~al.(2017)Jang, Gu, and Poole}]{jang2017categorical}
Eric Jang, Shixiang Gu, and Ben Poole. 2017.
\newblock Categorical reparametrization with gumble-softmax.
\newblock In \emph{International Conference on Learning Representations (ICLR 2017)}. OpenReview. net.

\bibitem[{Jing et~al.()Jing, Vincent, LeCun, and Tian}]{jingunderstanding}
Li~Jing, Pascal Vincent, Yann LeCun, and Yuandong Tian.
\newblock Understanding dimensional collapse in contrastive self-supervised learning.
\newblock In \emph{International Conference on Learning Representations}.

\bibitem[{Kolasani(2023)}]{kolasani2023optimizing}
Saydulu Kolasani. 2023.
\newblock Optimizing natural language processing, large language models (llms) for efficient customer service, and hyper-personalization to enable sustainable growth and revenue.
\newblock \emph{Transactions on Latest Trends in Artificial Intelligence}, 4(4).

\bibitem[{Kolodner(1992)}]{kolodner1992introduction}
Janet~L Kolodner. 1992.
\newblock An introduction to case-based reasoning.
\newblock \emph{Artificial intelligence review}, 6(1):3--34.

\bibitem[{Lester et~al.(2021)Lester, Al-Rfou, and Constant}]{lester2021power}
Brian Lester, Rami Al-Rfou, and Noah Constant. 2021.
\newblock The power of scale for parameter-efficient prompt tuning.
\newblock In \emph{Proceedings of the 2021 Conference on Empirical Methods in Natural Language Processing}, pages 3045--3059.

\bibitem[{Li et~al.(2020)Li, Li, Wang, Fu, Lin, Chen, Zhang, Tao, Zhang, Wang et~al.}]{li2020improving}
Jianqiao Li, Chunyuan Li, Guoyin Wang, Hao Fu, Yuhchen Lin, Liqun Chen, Yizhe Zhang, Chenyang Tao, Ruiyi Zhang, Wenlin Wang, et~al. 2020.
\newblock Improving text generation with student-forcing optimal transport.
\newblock In \emph{Proceedings of the 2020 Conference on Empirical Methods in Natural Language Processing (EMNLP)}, pages 9144--9156.

\bibitem[{Li and Liang(2021)}]{li2021prefix}
Xiang~Lisa Li and Percy Liang. 2021.
\newblock Prefix-tuning: Optimizing continuous prompts for generation.
\newblock In \emph{Proceedings of the 59th Annual Meeting of the Association for Computational Linguistics and the 11th International Joint Conference on Natural Language Processing (Volume 1: Long Papers)}, pages 4582--4597.

\bibitem[{Liu et~al.(2020)Liu, Li, and Sun}]{liu2020self}
Songtao Liu, Zeming Li, and Jian Sun. 2020.
\newblock Self-emd: Self-supervised object detection without imagenet.
\newblock \emph{arXiv preprint arXiv:2011.13677}.

\bibitem[{Liu et~al.(2022)Liu, Ji, Fu, Tam, Du, Yang, and Tang}]{liu2022ptuning}
Xiao Liu, Kaixuan Ji, Yicheng Fu, Weng Tam, Zhengxiao Du, Zhilin Yang, and Jie Tang. 2022.
\newblock P-tuning: Prompt tuning can be comparable to fine-tuning across scales and tasks.
\newblock In \emph{Proceedings of the 60th Annual Meeting of the Association for Computational Linguistics (Volume 2: Short Papers)}, pages 61--68.

\bibitem[{Madaan et~al.(2024)Madaan, Tandon, Gupta, Hallinan, Gao, Wiegreffe, Alon, Dziri, Prabhumoye, Yang et~al.}]{madaan2024self}
Aman Madaan, Niket Tandon, Prakhar Gupta, Skyler Hallinan, Luyu Gao, Sarah Wiegreffe, Uri Alon, Nouha Dziri, Shrimai Prabhumoye, Yiming Yang, et~al. 2024.
\newblock Self-refine: Iterative refinement with self-feedback.
\newblock \emph{Advances in Neural Information Processing Systems}, 36.

\bibitem[{Pan et~al.(2023)Pan, Saxon, Xu, Nathani, Wang, and Wang}]{pan2023automatically}
Liangming Pan, Michael Saxon, Wenda Xu, Deepak Nathani, Xinyi Wang, and William~Yang Wang. 2023.
\newblock Automatically correcting large language models: Surveying the landscape of diverse self-correction strategies.
\newblock \emph{arXiv preprint arXiv:2308.03188}.

\bibitem[{Pope et~al.(2023)Pope, Douglas, Chowdhery, Devlin, Bradbury, Heek, Xiao, Agrawal, and Dean}]{pope2023efficiently}
Reiner Pope, Sholto Douglas, Aakanksha Chowdhery, Jacob Devlin, James Bradbury, Jonathan Heek, Kefan Xiao, Shivani Agrawal, and Jeff Dean. 2023.
\newblock Efficiently scaling transformer inference.
\newblock \emph{Proceedings of Machine Learning and Systems}, 5:606--624.

\bibitem[{Pu and Demberg(2023)}]{pu2023chatgpt}
Dongqi Pu and Vera Demberg. 2023.
\newblock Chatgpt vs human-authored text: Insights into controllable text summarization and sentence style transfer.
\newblock In \emph{Proceedings of the 61st Annual Meeting of the Association for Computational Linguistics (Volume 4: Student Research Workshop)}, pages 1--18.

\bibitem[{Rawte et~al.(2023)Rawte, Sheth, and Das}]{rawte2023survey}
Vipula Rawte, Amit Sheth, and Amitava Das. 2023.
\newblock A survey of hallucination in large foundation models.
\newblock \emph{arXiv preprint arXiv:2309.05922}.

\bibitem[{Renze and Guven(2024)}]{renze2024self}
Matthew Renze and Erhan Guven. 2024.
\newblock Self-reflection in llm agents: Effects on problem-solving performance.
\newblock \emph{arXiv preprint arXiv:2405.06682}.

\bibitem[{Roziere et~al.(2023)Roziere, Gehring, Gloeckle, Sootla, Gat, Tan, Adi, Liu, Sauvestre, Remez et~al.}]{roziere2023code}
Baptiste Roziere, Jonas Gehring, Fabian Gloeckle, Sten Sootla, Itai Gat, Xiaoqing~Ellen Tan, Yossi Adi, Jingyu Liu, Romain Sauvestre, Tal Remez, et~al. 2023.
\newblock Code llama: Open foundation models for code.
\newblock \emph{arXiv preprint arXiv:2308.12950}.

\bibitem[{Rubner et~al.(2000)Rubner, Tomasi, and Guibas}]{rubner2000earth}
Yossi Rubner, Carlo Tomasi, and Leonidas~J Guibas. 2000.
\newblock The earth mover's distance as a metric for image retrieval.
\newblock \emph{International journal of computer vision}, 40:99--121.

\bibitem[{Shinn et~al.(2023)Shinn, Cassano, Berman, Gopinath, Narasimhan, and Yao}]{shinn2023reflexion}
Noah Shinn, Federico Cassano, Edward Berman, Ashwin Gopinath, Karthik Narasimhan, and Shunyu Yao. 2023.
\newblock \href {https://arxiv.org/abs/2303.11366} {Reflexion: Language agents with verbal reinforcement learning}.
\newblock \emph{Preprint}, arXiv:2303.11366.

\bibitem[{Turpin et~al.(2024)Turpin, Michael, Perez, and Bowman}]{turpin2024language}
Miles Turpin, Julian Michael, Ethan Perez, and Samuel Bowman. 2024.
\newblock Language models don't always say what they think: unfaithful explanations in chain-of-thought prompting.
\newblock \emph{Advances in Neural Information Processing Systems}, 36.

\bibitem[{Wei et~al.(2022{\natexlab{a}})Wei, Tay, Bommasani, Raffel, Zoph, Borgeaud, Yogatama, Bosma, Zhou, Metzler et~al.}]{wei2022emergent}
Jason Wei, Yi~Tay, Rishi Bommasani, Colin Raffel, Barret Zoph, Sebastian Borgeaud, Dani Yogatama, Maarten Bosma, Denny Zhou, Donald Metzler, et~al. 2022{\natexlab{a}}.
\newblock Emergent abilities of large language models.
\newblock \emph{arXiv preprint arXiv:2206.07682}.

\bibitem[{Wei et~al.(2022{\natexlab{b}})Wei, Wang, Schuurmans, Bosma, Xia, Chi, Le, Zhou et~al.}]{wei2022chain}
Jason Wei, Xuezhi Wang, Dale Schuurmans, Maarten Bosma, Fei Xia, Ed~Chi, Quoc~V Le, Denny Zhou, et~al. 2022{\natexlab{b}}.
\newblock Chain-of-thought prompting elicits reasoning in large language models.
\newblock \emph{Advances in neural information processing systems}, 35:24824--24837.

\bibitem[{Xin et~al.(2020)Xin, Tang, Lee, Yu, and Lin}]{2020deebert}
Ji~Xin, Raphael Tang, Jaejun Lee, Yaoliang Yu, and Jimmy Lin. 2020.
\newblock \href {https://doi.org/10.18653/v1/2020.acl-main.204} {{D}ee{BERT}: Dynamic early exiting for accelerating {BERT} inference}.
\newblock In \emph{Proceedings of the 58th Annual Meeting of the Association for Computational Linguistics}, pages 2246--2251, Online. Association for Computational Linguistics.

\bibitem[{Yang et~al.(2024)Yang, Yang, Hui, Zheng, Yu, Zhou, Li, Li, Liu, Huang et~al.}]{yang2024qwen2}
An~Yang, Baosong Yang, Binyuan Hui, Bo~Zheng, Bowen Yu, Chang Zhou, Chengpeng Li, Chengyuan Li, Dayiheng Liu, Fei Huang, et~al. 2024.
\newblock Qwen2 technical report.
\newblock \emph{arXiv preprint arXiv:2407.10671}.

\bibitem[{Yao et~al.(2024)Yao, Yu, Zhao, Shafran, Griffiths, Cao, and Narasimhan}]{yao2024tree}
Shunyu Yao, Dian Yu, Jeffrey Zhao, Izhak Shafran, Tom Griffiths, Yuan Cao, and Karthik Narasimhan. 2024.
\newblock Tree of thoughts: Deliberate problem solving with large language models.
\newblock \emph{Advances in Neural Information Processing Systems}, 36.

\bibitem[{Zhang et~al.(2020)Zhang, Cai, Lin, and Shen}]{Zhang_2020_CVPR}
Chi Zhang, Yujun Cai, Guosheng Lin, and Chunhua Shen. 2020.
\newblock Deepemd: Few-shot image classification with differentiable earth mover's distance and structured classifiers.
\newblock In \emph{IEEE/CVF Conference on Computer Vision and Pattern Recognition (CVPR)}.

\bibitem[{Zhang et~al.(2024{\natexlab{a}})Zhang, Wu, Xu, Cao, Du, and Psounis}]{zhang2024efficient}
Jiang Zhang, Qiong Wu, Yiming Xu, Cheng Cao, Zheng Du, and Konstantinos Psounis. 2024{\natexlab{a}}.
\newblock Efficient toxic content detection by bootstrapping and distilling large language models.
\newblock In \emph{Proceedings of the AAAI Conference on Artificial Intelligence}, volume~38, pages 21779--21787.

\bibitem[{Zhang et~al.(2023)Zhang, Han, Liu, Gao, Zhou, Hu, Yan, Lu, Li, and Qiao}]{zhang2023llama}
Renrui Zhang, Jiaming Han, Chris Liu, Peng Gao, Aojun Zhou, Xiangfei Hu, Shilin Yan, Pan Lu, Hongsheng Li, and Yu~Qiao. 2023.
\newblock Llama-adapter: Efficient fine-tuning of language models with zero-init attention.
\newblock \emph{arXiv preprint arXiv:2303.16199}.

\bibitem[{Zhang et~al.(2024{\natexlab{b}})Zhang, Lin, Liu, Shu, Li, Zhang, Wanggui, Zhou, Lv, Jiang et~al.}]{zhang2024hyperllava}
Wenqiao Zhang, Tianwei Lin, Jiang Liu, Fangxun Shu, Haoyuan Li, Lei Zhang, He~Wanggui, Hao Zhou, Zheqi Lv, Hao Jiang, et~al. 2024{\natexlab{b}}.
\newblock Hyperllava: Dynamic visual and language expert tuning for multimodal large language models.
\newblock \emph{arXiv preprint arXiv:2403.13447}.

\bibitem[{Zhu et~al.(2022)Zhu, Guo, Wu, and Tang}]{rosa}
Yun Zhu, Jianhao Guo, Fei Wu, and Siliang Tang. 2022.
\newblock Rosa: A robust self-aligned framework for node-node graph contrastive learning.
\newblock \emph{arXiv preprint arXiv:2204.13846}.

\end{thebibliography}

\appendix
\clearpage
\section{Sinkhorn Algorithm and Optimal Transport}~\label{app:appro_algo}
The vanilla optimization problem of optimal transport, as formulated in Equation~\ref{equ:ot}, aims to find the optimal transportation matrix~\( \tilde{\mathbf{\Gamma}}\). Nevertheless, the exact minimization over ~\( \tilde{\mathbf{\Gamma}}\) is generally computationally intractable~\cite{arjovsky2017wasserstein,genevay2018learning,li2020improving}. To address this, the~\emph{Sinkhorn Algorithm}~\cite{Cuturi_2013} is utilized to approximate~\( \tilde{\mathbf{\Gamma}}\). Specifically, the algorithm introduces a regularization term:
\begin{equation}
    \min\limits_{\mathbf{\Gamma \in \Pi(\mathbf{r},\mathbf{c})}} \langle \mathbf{\Gamma}, \mathbf{D}\rangle_\mathbf{\mathrm{F}} + \underbrace{\frac{1}{\lambda}\mathbf{\Gamma}(\log {\mathbf{\Gamma}}-1)}_{\text{regularization term}},
\end{equation}
where \(\langle , \rangle_\mathrm{F}\) denotes Frobenius inner product, and \(\lambda\) is a hyper-parameter that controls the strength of regularization. 

With this regularization term, the optimal~\(\tilde{\Gamma}\) can be approximated as:
\begin{equation}
    \tilde{\mathbf{\Gamma}} = diag(\boldsymbol{v})\mathbf{Q} diag(\boldsymbol{u}),
\end{equation}
where \(\mathbf{Q}=e^{-\lambda \mathbf{D}}\), and  \(\boldsymbol{v}\), \(\boldsymbol{u}\) are two coefficient vectors whose values can be iteratively updated as:
\begin{align}
    & \boldsymbol{v}^{t+1}_i = \frac{\boldsymbol{r}_i}{\sum^{k}_{j=1}{\mathbf{Q}_{ij}\boldsymbol{u}^t_j}},\notag \\
    & \boldsymbol{u}^{t+1}_j = \frac{\boldsymbol{c}_j}{\sum^{k^{\prime}}_{i=1}{\mathbf{Q}_{ij}\boldsymbol{v}^{t+1}_i}}
\end{align}
The critical aspect then lies in determining the marginal weights~\(\boldsymbol{r}\) and~\(\boldsymbol{c}\), which control the total supplying and demanding units, respectively. A larger weight indicates that the reflective unit exhibits semantic similarity to the ground truth reflection tokens. We define the weight as dot product between its embedding and the mean pooling embedding from the other set:
\begin{align}
    & \boldsymbol{r}_i = \max\{{{\boldsymbol{p}^{\prime}}^T_i \cdot \frac{\sum^k_{j=1}{\boldsymbol{p}_j}}{k},0}\}, \notag \\
    & \boldsymbol{c}_j = \max\{{\boldsymbol{p}^T_j \cdot \frac{\sum^{k^{\prime}}_{i=1}{{\boldsymbol{p}^{\prime}}_i}}{k^{\prime}},0}\}
\end{align}

After obtaining the approximated optimal transportation matrix \(\tilde{\Gamma}\), we can compute the loss as defined in Equation~\ref{equ:loss}.

\section{E-commerce Customer Intent Detection (ECID) Benchmark}~\label{app:ecid}
In the domain of Intelligent Customer Service (ICS) for e-commerce, effectively and efficiently discerning customers' core intentions when they contact ICS for assistance is critical to enhancing service quality~\cite{cheng2024impact, kolasani2023optimizing}. In this work, we introduce an industrial benchmark, named E-commerce Customer Intent Detection (ECID) to evaluate our proposed method. This dataset is in Chinese, focusing on customer interactions within major Chinese e-commerce platforms. The following sections detail the construction of this dataset and elaborate on its specific tasks.

\paragraph{Task.}
The primary objective of the ECID dataset is to infer the core intention of customers seeking ICS assistance, based on previous communication records between customers and customer service platforms, customer purchase histories, and order information. The core intention refers to the customer's current concern or the problem they wish to resolve.
Specifically, each data point in the dataset comprises input information from five fields:
\begin{itemize}
    \item  \textbf{Customer Question.} The specific issue or obstacle encountered by the customer.
    \item  \textbf{Customer Request.} Customer requirements, encompassing all objectives or desired outcomes expressed during interactions with the ICS, sellers, and platform customer service representatives, as well as any proactively initiated request.
    \item \textbf{Solution.} Proposals offered by the platform or sellers to address the customer's issue.
    \item \textbf{Customer Attitude.} The customer's attitudes towards the proposed solutions, as expressed during communication.
    \item \textbf{Processing status.} PThe current state of the customer's submitted request.
\end{itemize}

\begin{figure}
    \centering
    \includegraphics[width=1\linewidth]{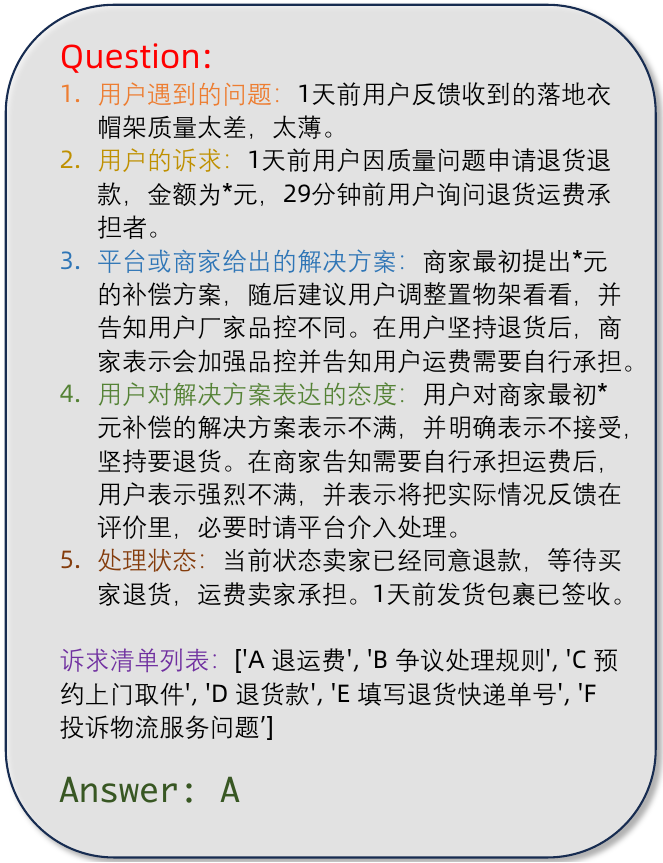}
   \caption{An example of the ECID dataset.}
   \label{fig:ecid_case}
    \vspace{-0.5em}
\end{figure}

ECID aims to match the aforementioned input information with the most appropriate intention from a predefined list. In real-world applications, we categorize intentions into 36 distinct types, each representing a specific issue customers seek to resolve. For the ECID dataset, a condensed list of six intentions is provided, from which the most relevant core intention must be selected. An illustrative example is presented in the accompanying Figure~\ref{fig:ecid_case}.

\paragraph{Data Processing.}
The ECID dataset is derived from customer service system records of the Taobao e-commerce platform, collected over a single day. From this collection, we randomly sampled 30,000 data points in an unbiased manner. Each data point comprises information from various sources, including customer-service representative chat logs, customer-seller communications, customer order details, and ongoing request processing records. We employed a fine-tuned LLM, specifically \texttt{Qwen2-7B-Instruct}, to extract the aforementioned five fields of information from the diverse sources.

We initially applied a rule-based method to eliminate incomplete or inconsistent data (such as newly registered users without any purchase history), resulting in approximately 4,000 refined data points. Subsequently, we utilized \texttt{GPT-4-turbo-128k} and \texttt{Qwen2-72B-Instruct} for data labeling. Using a voting system, we selected the most appropriate intention from a predefined list of 36 intentions, along with five secondary matching intentions, to create a set of candidate intentions and answers for each data point.
To ensure high data quality, we discarded instances where the highest voting rate was below 80\%. We also implemented human evaluation, randomly sampling and verifying the accuracy of answers. This rigorous process yielded \textbf{1,170} high-quality data points, each accompanied by a Chain-of-Thought (CoT) reasoning process. The dataset was partitioned into a 7:3 ratio for training and testing.

We conducted data anonymization to remove sensitive information from the dataset. Personal identifiable information, including customer names, addresses, and contact details, was redacted. Additionally, all monetary values within the dataset were masked using asterisks (*) to ensure confidentiality.

\section{Public Datasets}~\label{app:datasets}
We evaluate our method across three public datasets spanning diverse domains: two programming benchmarks (MBPP and HumanEval) and one mathematical reasoning dataset (GSM8K).

\paragraph{Programming.}
For evaluating our method on programming tasks, we utilize two Python code programming benchmarks: MBPP~\cite{austin2021program} and HumanEval~\cite{chen2021evaluating}. The MBPP dataset consists of approximately 1,000 Python programming problems, while HumanEval encompasses 161 problems, each accompanied by comprehensive unit test cases. We adhere to the official train-test split for MBPP, employing its training set for model training. As HumanEval provides only a test set, we use it exclusively for evaluation purposes. Following~\citet{dou2024reflection}, we employ the Pass@k metric, which quantifies the percentage of problems where the model successfully passes all unit tests within k attempts. During the code generation process, in line with previous work by~\citet{roziere2023code}, the actor model is provided with the unit test cases.

\paragraph{Mathematical Reasoning.}
For mathematical reasoning evaluation, we employ the Grade School Math 8K (GSM8K) dataset~\cite{cobbe2021gsm8k}, a comprehensive benchmark containing approximately 8,000 grade school mathematics word problems. This dataset is particularly valuable due to its linguistic diversity and high-quality annotations, featuring detailed human-curated solution trajectories and precise answers for each problem~\cite{madaan2024self}. Following standard practices, we strictly adhere to the official train-test split (7,473 for training, 1,319 for testing) in our experimental setup. Performance is evaluated using the Exact Match (EM) metric, which assesses the precise correspondence between model-generated responses and ground-truth answers~\cite{madaan2024self}, providing a rigorous measure of mathematical reasoning capabilities.

\section{Baselines}\label{app:baseline}
We evaluate our method against three categories of baselines: \textbf{Common Reasoning}, \textbf{Parameter-Efficient Fine-Tuning (PEFT)}, and \textbf{Reflection-Based} approaches. The specifics of these baseline implementations are detailed below:

\paragraph{Common Reasoning Approaches.}
For common reasoning approaches, we evaluate both \textbf{Zero-Shot} and \textbf{Few-Shot} (2-shots)~\cite{fewshot} strategies. In both settings, we employ the Chain-of-Thought (CoT)~\cite{wei2022chain} reasoning methodology to facilitate structured generation processes.

\paragraph{Parameter-Efficient Fine-Tuning (PEFT) Approaches.}
We implement three widely-adopted PEFT methods for model tuning: \textbf{LoRA}~\cite{hu2021lora}, \textbf{P-Tuning}~\cite{liu2022ptuning}, and \textbf{Llama-Adapter}~\cite{zhang2023llama}. Through extensive hyper-parameter grid search:
For LoRA, we augment the query, key, and value matrices with adapter matrices of rank \{8, 16\}.
For P-Tuning, we experiment with prompt lengths of \{16, 32, 64\} and implement the MLP-based re-parameterization function~\cite{liu2022ptuning}.
For Llama-Adapter, we explore adapter lengths of \{32, 64\} and position them within the final 15 layers of the LLM~\cite{zhang2023llama}.
 
\paragraph{Reflection-Based Approaches.}
We implement two reflection-based approaches as our primary baselines:

\textbf{Re-ReST}~\cite{dou2024reflection} implements a self-reflection mechanism to optimize self-training data quality. The method operates in two phases: first refining the training dataset through reflective incorporation, then conducting model fine-tuning on the enhanced data. This approach enables implicit integration of reflective insights, allowing for improved performance during single-pass inference. We employ their official implementation\footnote{https://github.com/PlusLabNLP/Re-ReST}, adapting it to our experimental settings with corresponding datasets and base LLMs.

\textbf{Reflection-RAG} implements a Retrieval Augmented Generation (RAG) framework~\cite{gao2023retrieval} for reflection-based reasoning. The method stores reflections generated from the training dataset as described in Section~\ref{method:ref_gen}. During inference, it retrieves relevant reflections based on question similarity, leveraging the intuition that similar questions often share comparable solution strategies and hints.
The retrieval process consists of two phases for enhanced accuracy:
First, we employ \texttt{BGE-m3}\footnote{https://huggingface.co/BAAI/bge-m3}, a widely-adopted text embedding model for RAG systems, to identify the top-6 similar question-reflection pairs.
Subsequently, we utilize \texttt{BGE-reranker-v2-m3}\footnote{https://huggingface.co/BAAI/bge-reranker-v2-m3} to re-rank these candidates and select the reflection whose associated question exhibits the highest relevance to the input query.
The selected reflection then serves as guidance for the LLM's problem-solving process. To optimize retrieval efficiency, we cache question embedding matrices in GPU memory, significantly reducing retrieval latency.

\section{Implementations Details}
\paragraph{Models.}
To evaluate our proposed approach, we employ three widely used base LLMs as Actor LLMs: \texttt{Qwen-2-7B-Instruct}\footnote{https://huggingface.co/Qwen/Qwen2-7B-Instruct}~\cite{yang2024qwen2}, \texttt{Llama-3.1-8B-Instruct}\footnote{https://huggingface.co/meta-llama/Llama-3.1-8B-Instruct}~\cite{dubey2024llama}, and \texttt{CodeLlama-7B-Instruct}\footnote{https://huggingface.co/meta-llama/CodeLlama-7b-Instruct-hf}~\cite{roziere2023code}. Additionally, we utilize \texttt{Qwen-2-72B-Instruct}\footnote{https://huggingface.co/Qwen/Qwen2-72B-Instruct} as the Reflector Model in our experiments.

\paragraph{Implementations Details.}
In the reflection generation phase, we set a maximum of 4 iteration steps, discarding data that fails to solve the problem correctly after 4 action-reflection loops. To ensure certainty, we set the reflector LLM's temperature to 0, eliminating sampling variability.

For codebook tuning, we employ grid search to identify optimal hyper-parameters across various tasks. The codebook size is selected from \{512, 1024\}, positioned at either the last 3rd, 6th, or 9th layer. The number of selected reflective units is chosen from \{16, 32, 64\}.

We implement a progressive optimization paradigm to enhance model performance. During meta-reflection alignment, we set the epoch to either 1 or 2 with a learning rate of 1e-4. We utilize the Sinkhorn Algorithm to approximate the transportation matrix, with 10 iterations to ensure accurate approximation (details in Appendix~\ref{app:appro_algo}).

For supervised fine-tuning (SFT), we explore either 2 or 3 epochs with learning rates selected from \{1e-4, 5e-5, 1e-5\} for tuning.

\section{Prompts}
In this section, we present the domain-specific prompt templates utilized in our approach for various task domains. We emphasize that the `\textcolor[rgb]{0,0,0.8}{\{reflection\}}' component is only integrated into the prompt after the actor LLMs' first attempt. Initial trials are executed without any reflective guidance to establish baseline performance.

\subsection{Prompts for Programming Tasks}

\begin{tcolorbox}[
  breakable,
  colback=gray!20,
  colframe=black,
  boxrule=1pt
]
\setlength{\parindent}{0pt} 
\textbf{Prompt for Actor LLMs:}

You are an AI that only responds with python code, NOT ENGLISH. You will be given a function signature and its docstring by the user. Write your full implementation (restate the function signature, the class definition, or the necessary libraries).

[Function signature]: \textcolor[rgb]{0.8,0,0}{\{func\_sign\}}

[Your code should pass these tests]: \textcolor[rgb]{0,0.7,0}{\{unit tests\}}

[Hint or past experience that may guide you]:  \textcolor[rgb]{0,0,0.8}{\{reflection\}}

\end{tcolorbox}

\setlength{\parindent}{0pt} 
\begin{tcolorbox}[
  breakable,
  colback=gray!20,
  colframe=black,
  boxrule=1pt
]
\textbf{Prompt for Reflector LLMs:}

You are a Python programming assistant, your task is to instruct a student on correcting a mistake in a programming question. You will be given:
 
 1. A function signature.
 
 2. The student's implementation
 
 3. A series of unit tests for the implementation. 
 
Your goal is to write a few sentences to provide a corrective solution that can solve not only this question but also a series of similar questions. Remember point out the common pitfalls or easily misunderstood aspects of this problem based on the student's incorrect implementation. Then the student need this as a hint when he/she try again later. Only provide the few sentence description in your answer, not the implementation. 

Example output: `The hint to this programming problem is ...'

[Function signature]: \textcolor[rgb]{0.8,0,0}{\{func\_sign\}}

[Function impl]: \textcolor[rgb]{0,0.8,0.8}{\{fun\_impl\}}

[Unit test results]:  \textcolor[rgb]{0.8,0,0.8}{\{test results\}}

\end{tcolorbox}

\subsection{Prompts for Mathematical Reasoning Task}
\setlength{\parindent}{0pt} 
\begin{tcolorbox}[
  breakable,
  colback=gray!20,
  colframe=black,
  boxrule=1pt
]
\textbf{Prompt for Actor LLMs:}

You are an AI assitant, you are required to solve mathematical question.

[Question]: \textcolor[rgb]{0.8,0,0}{\{question\}}

[Hint or past experience that may guide you]:  \textcolor[rgb]{0,0,0.8}{\{reflection\}}

\end{tcolorbox}

\setlength{\parindent}{0pt} 
\begin{tcolorbox}[
  breakable,
  colback=gray!20,
  colframe=black,
  boxrule=1pt
]
\textbf{Prompt for Reflector LLMs:}

You are a mathematical expert, your task is to instruct a student on correcting \
a mistake in a math question. Note that you should ONLY provide a corrective solution \
that can solve not only this question but also a series of similar questions, and you \
must not reveal the answer to prevent leaking. Your output should only contain the solution \
without any explanation. 

Example output: ‘For this question, you should first calculate...`

[Question]: \textcolor[rgb]{0.8,0,0}{\{question\}}

[Student response]: \textcolor[rgb]{0,0.8,0.8}{\{response\}}

\end{tcolorbox}

\subsection{Prompts for E-commerce Customer Intent Detection Task}
\setlength{\parindent}{0pt} 
\begin{tcolorbox}[
  breakable,
  colback=gray!20,
  colframe=black,
  boxrule=1pt
]
\textbf{Prompt for Actor LLMs:}

\begin{CJK}{UTF8}{gbsn}
你是一个来自电商平台的AI客服智能助手，你的输入分为两部分：

\#\# 用户需求以及订单的信息，分为以下五个字段内容：
1. 用户遇到的问题，即用户遭遇到的异常情况或障碍；
2. 用户的诉求，即用户所有的在与助手、商家和平台人工客服沟通过程中表达的想要实现的目的或达成的内容以及主动发起的申请，包括退款申请、投诉申请、赔偿申请等；
3. 平台或商家给出的解决方案；
4. 用户对解决方案表达的态度；
5. 处理状态；

\#\# 定义好的诉求清单，用列表作为输入，其中一共有6个诉求，诉求由字母+诉求文字表示（比如 'B 退运费'）

\#\# 你现在需要根据以上信息从诉求清单列表中选择出最匹配的用户诉求，你的输出应该包括：
1.你的思考过程
2.诉求清单中最为匹配的诉求对应的字母，有且仅有一个。

[问题]：\textcolor[rgb]{0.8,0,0}{\{question\}}

[一些可能对你有用的提示和来自过去的错误经验]：\textcolor[rgb]{0,0,0.8}{\{reflection\}}
\end{CJK}
\end{tcolorbox}

\setlength{\parindent}{0pt} 
\begin{tcolorbox}[
  breakable,
  colback=gray!20,
  colframe=black,
  boxrule=1pt
]
\textbf{Prompt for Reflector LLMs:}

\begin{CJK}{UTF8}{gbsn}
你是一个智能AI助手，现在需要你解决一些电商智能助手在推断用户诉求时存在的问题。
目前输入分为三部分内容：

\#\# 用户需求以及订单的信息，分为以下五个字段内容：
1. 用户遇到的问题，即用户遭遇到的异常情况或障碍；
2. 用户的诉求，即用户所有的在与客服、商家和平台人工客服沟通过程中表达的想要实现的目的或达成的内容以及主动发起的申请，包括退款申请、投诉申请、赔偿申请等；
3. 平台或商家给出的解决方案；
4. 用户对解决方案表达的态度；
5. 处理状态；

\#\# 定义好的诉求清单，用列表作为输入，其中一共有6个诉求，诉求由字母+诉求文字表示（比如 'A 退运费'），核心任务是根据用户需求和订单信息选择出最匹配的诉求

\#\# 一段错误的匹配过程，其中包括思考过程和预测的诉求

现在需要你对上述错误的匹配过程的进行反思，并提供正确的解决方案，以指导再次遇到类似订单情况下能够找出最匹配的诉求。注意，你的输出不应该包括正确答案（防止出现答案泄漏），应该给出如何思考从而指导下一次的匹配过程，并且保证通用性（对相似问题也可以提供帮助）。'''

[问题]: \textcolor[rgb]{0.8,0,0}{\{question\}}

[匹配过程]: \textcolor[rgb]{0,0.8,0.8}{\{response\}}
\end{CJK}
\end{tcolorbox}

\section{Case Study}\label{app:case_study}
\begin{figure*}
    \centering
    \includegraphics[width=1\linewidth]{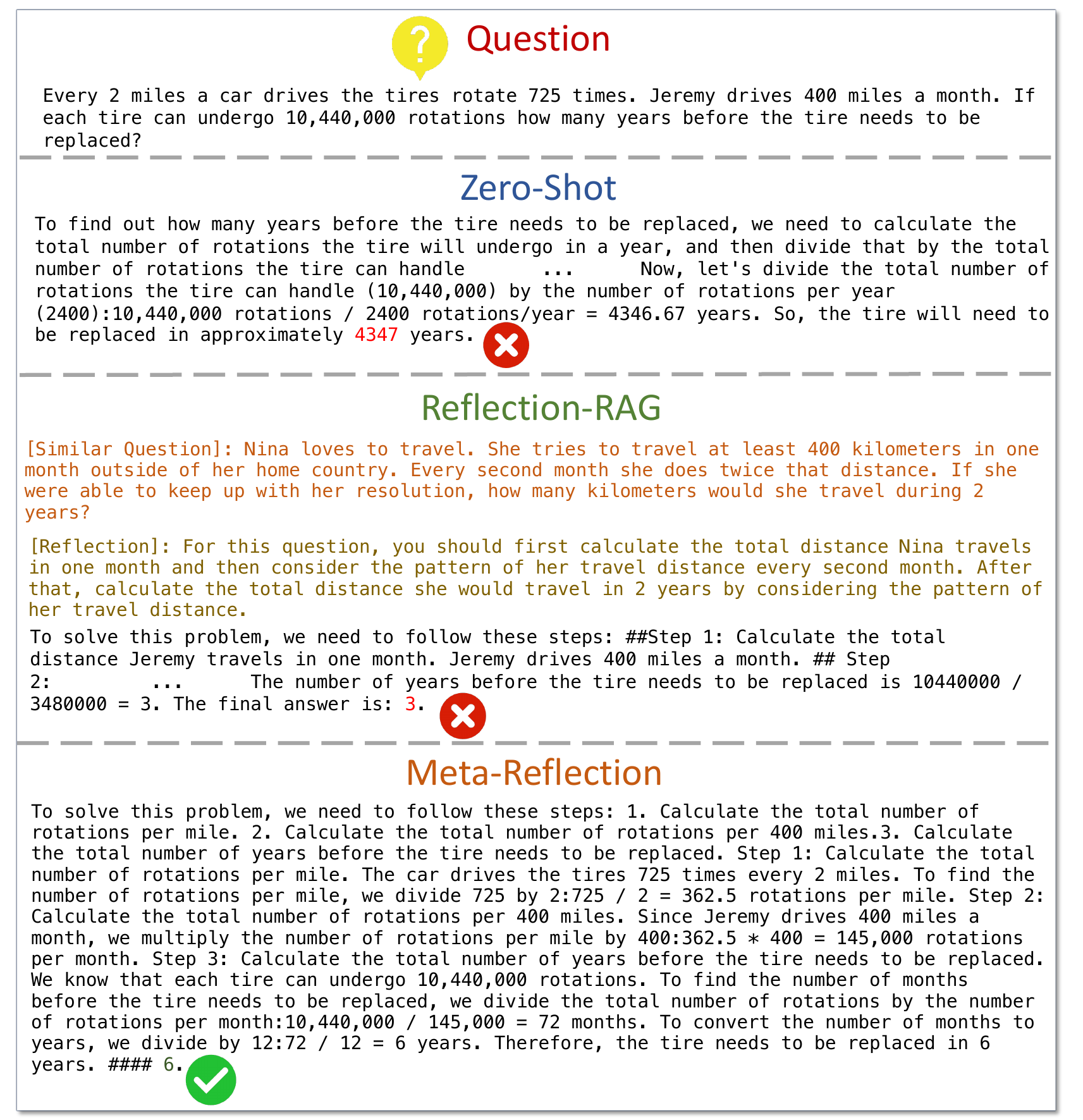}
   \caption{A case study on the GSM8K dataset.}\label{fig:case_study1}
\end{figure*}

We conduct a case study on the GSM8K dataset. As illustrated in Figure~\ref{fig:case_study1}, we compare three distinct methodologies. The base LLM, under Zero-Shot settings, demonstrates a lack of domain-relevant knowledge and fails to solve the problem without external guidance. The Reflection-RAG approach retrieves similar problems from the training knowledge base and leverages their associated reflections as guidance. However, despite the high similarity of retrieved problems, their reflection guidance processes often deviate significantly from the required reasoning path of the given problem. This misalignment prevents fine-grained guidance and introduces noise, resulting in suboptimal performance. In contrast, our proposed method achieves superior performance by incorporating reflective insights into the codebook and retrieving question-specific reflective insights during inference, enabling precise step-by-step guidance for the LLM to successfully solve the problem.

\section{Visualization}\label{app:vis}

\begin{figure*}[t]  
    \centering
    \begin{subfigure}[b]{0.48\textwidth} 
        \centering
        \includegraphics[width=\textwidth]{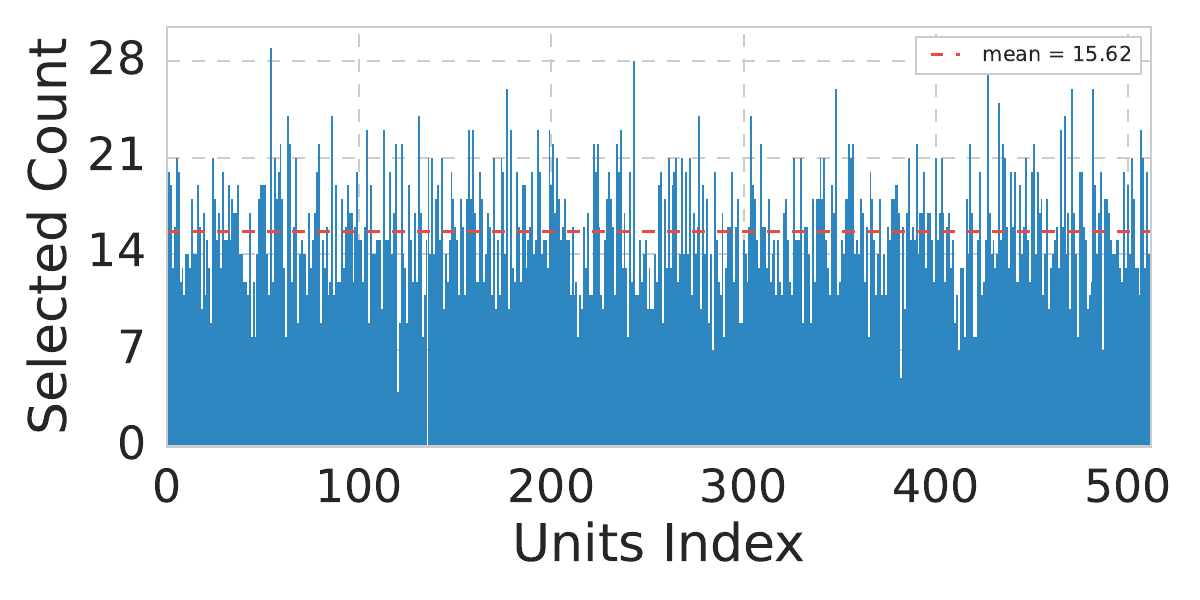}
        \caption{\texttt{MBPP}}
        \label{fig:units_distribution_mbpp}
    \end{subfigure}
    \hfill 
    \begin{subfigure}[b]{0.48\textwidth}
        \centering
        \includegraphics[width=\textwidth]{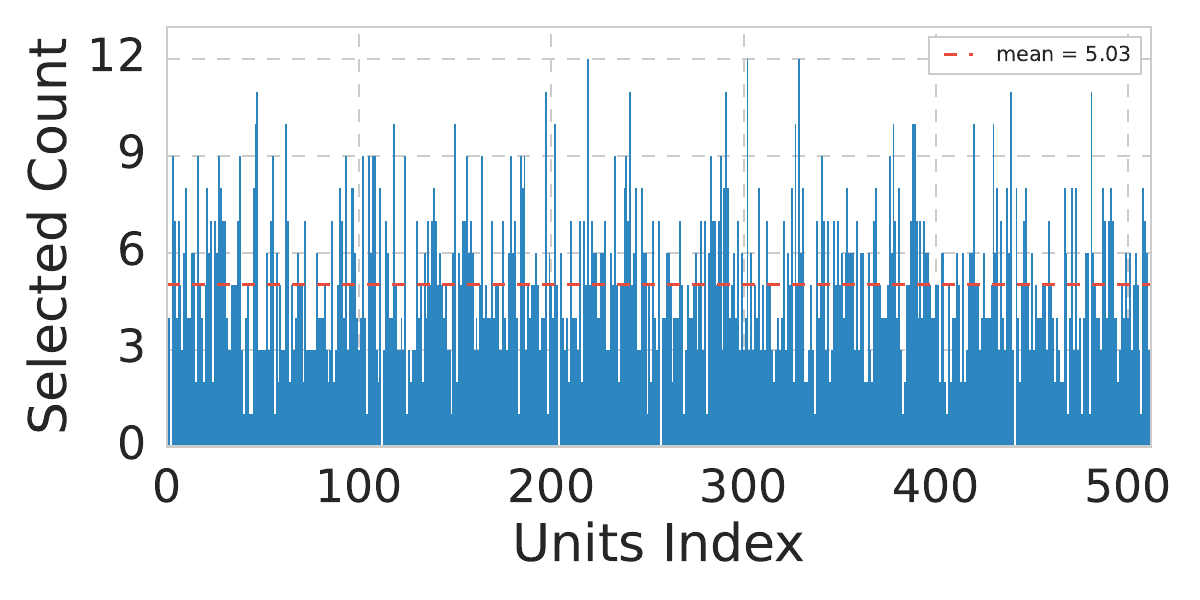}
        \caption{\texttt{HumanEval}}
        \label{fig:units_distribution_humaneval}
    \end{subfigure}
    
    \vspace{1em} 
    
    \begin{subfigure}[b]{0.48\textwidth}
        \centering
        \includegraphics[width=\textwidth]{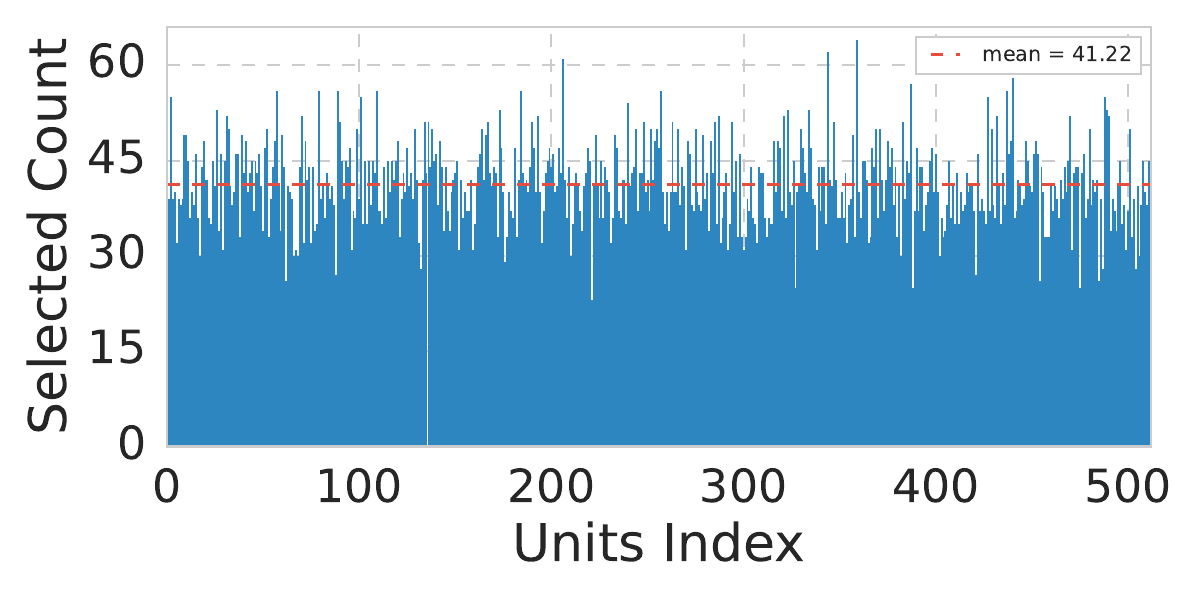}
        \caption{\texttt{GSM8K}}
        \label{fig:units_distribution_gsm8k}
    \end{subfigure}
    \caption{Visualization of reflective unit selection frequency distributions across three benchmark datasets in programming and mathematical reasoning domains. A meta-reflection codebook of size 512 is uniformly maintained, with 16 units uniformly selected per inference.}
    \label{fig:units_distribution_appendix}
\end{figure*}

We visualize the selection frequency distribution of reflective units in the meta-reflection codebook across three additional benchmark datasets from programming and mathematical reasoning domains. As shown in Figure~\ref{fig:units_distribution_appendix}, the reflective units exhibit significant variations, consistent with the findings in Section~\ref{exp:vis}. This distribution pattern indicates that the retrieval process adaptively selects different reflective units based on the specific questions, thereby providing tailored guidance for LLMs in problem-solving tasks.

\begin{figure*}[t]  
    \centering
    \begin{subfigure}[b]{0.48\textwidth} 
        \centering
        \includegraphics[width=\textwidth]
        {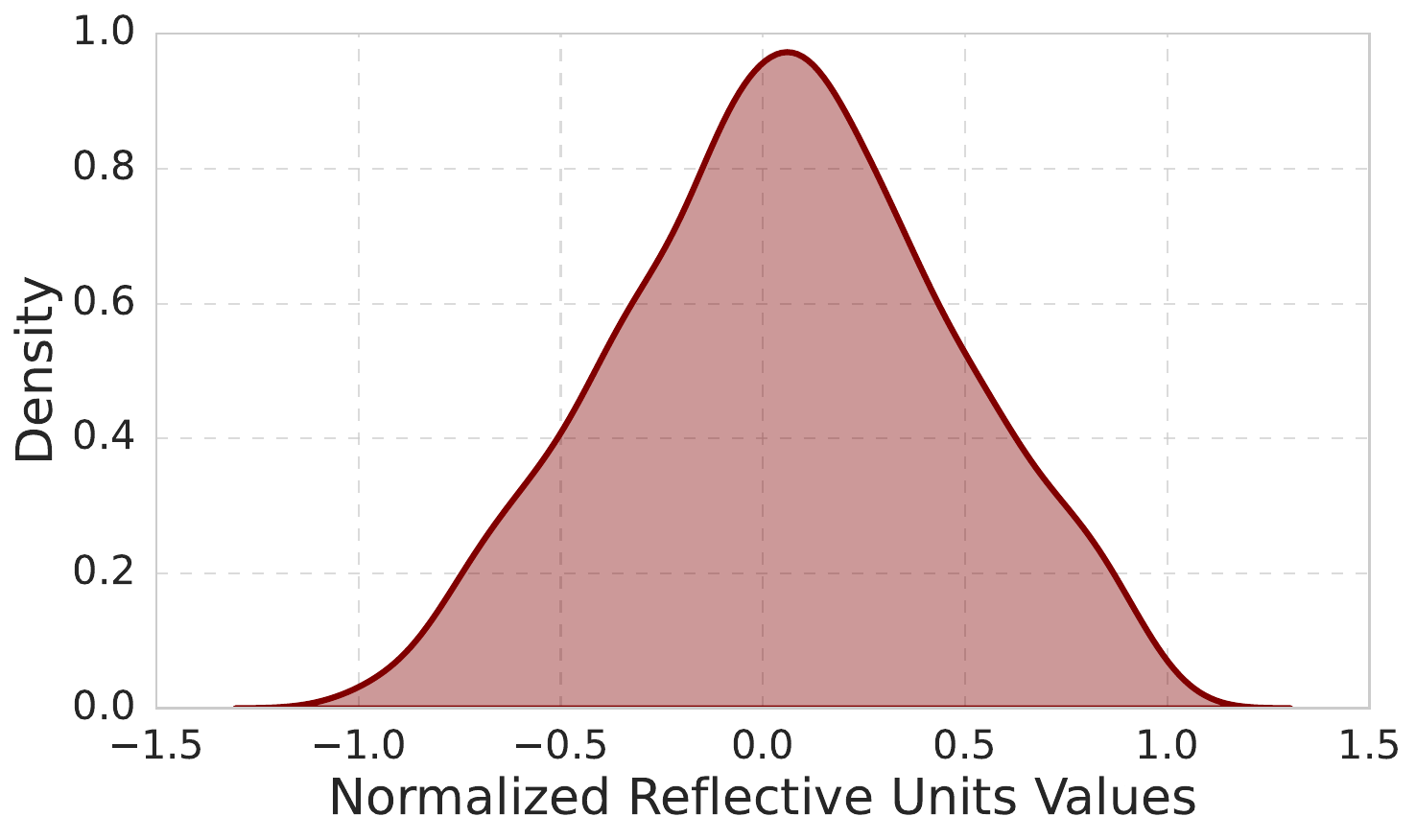}
        \caption{\texttt{MBPP}}
        \label{fig:feature_distribution_mbpp}
    \end{subfigure}
    \hfill 
    \begin{subfigure}[b]{0.48\textwidth}
        \centering
        \includegraphics[width=\textwidth]{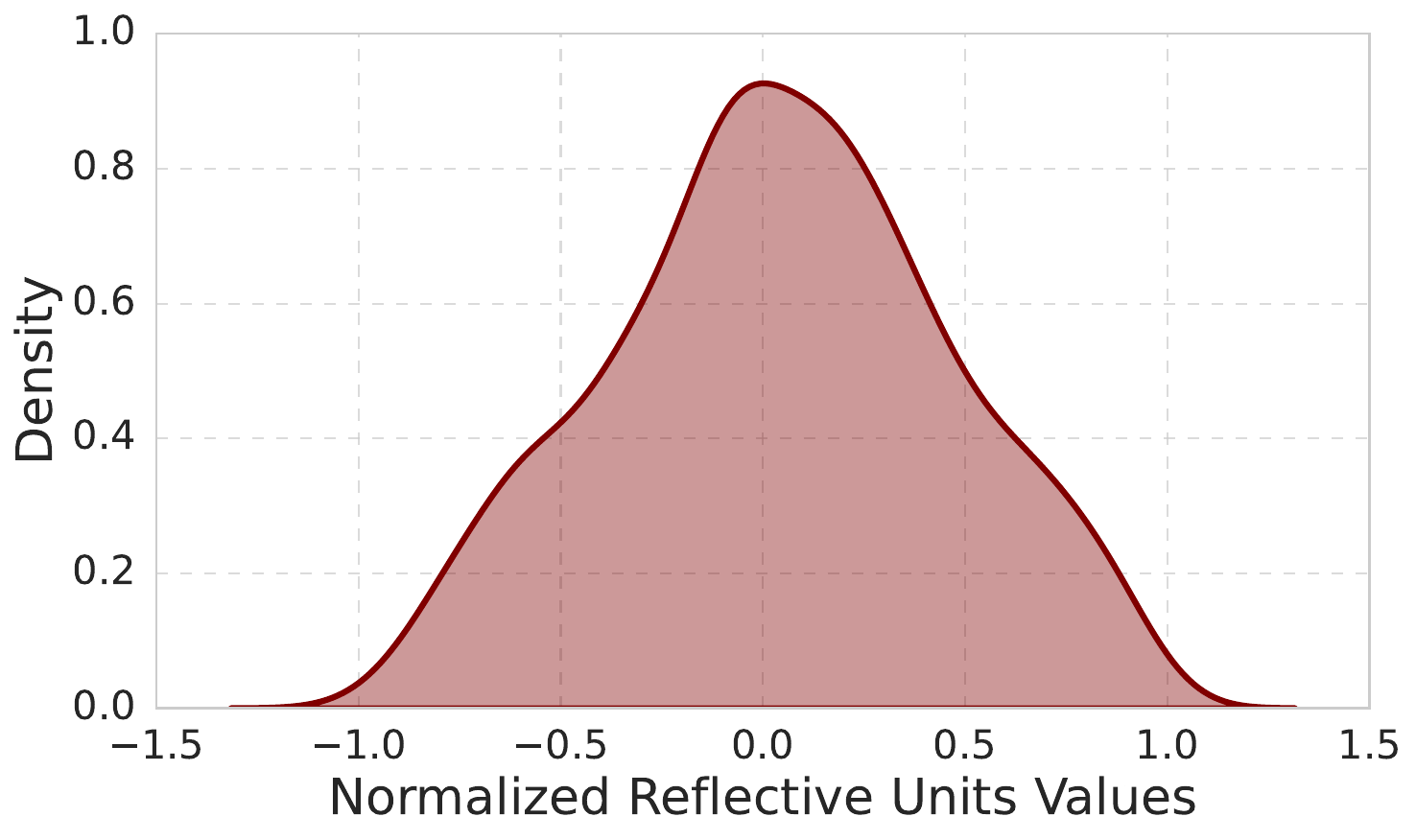}
        \caption{\texttt{HumanEval}}
        \label{fig:feature_distribution_humaneval}
    \end{subfigure}
    
    \vspace{1em} 
    
    \begin{subfigure}[b]{0.48\textwidth}
        \centering
        \includegraphics[width=\textwidth]
        {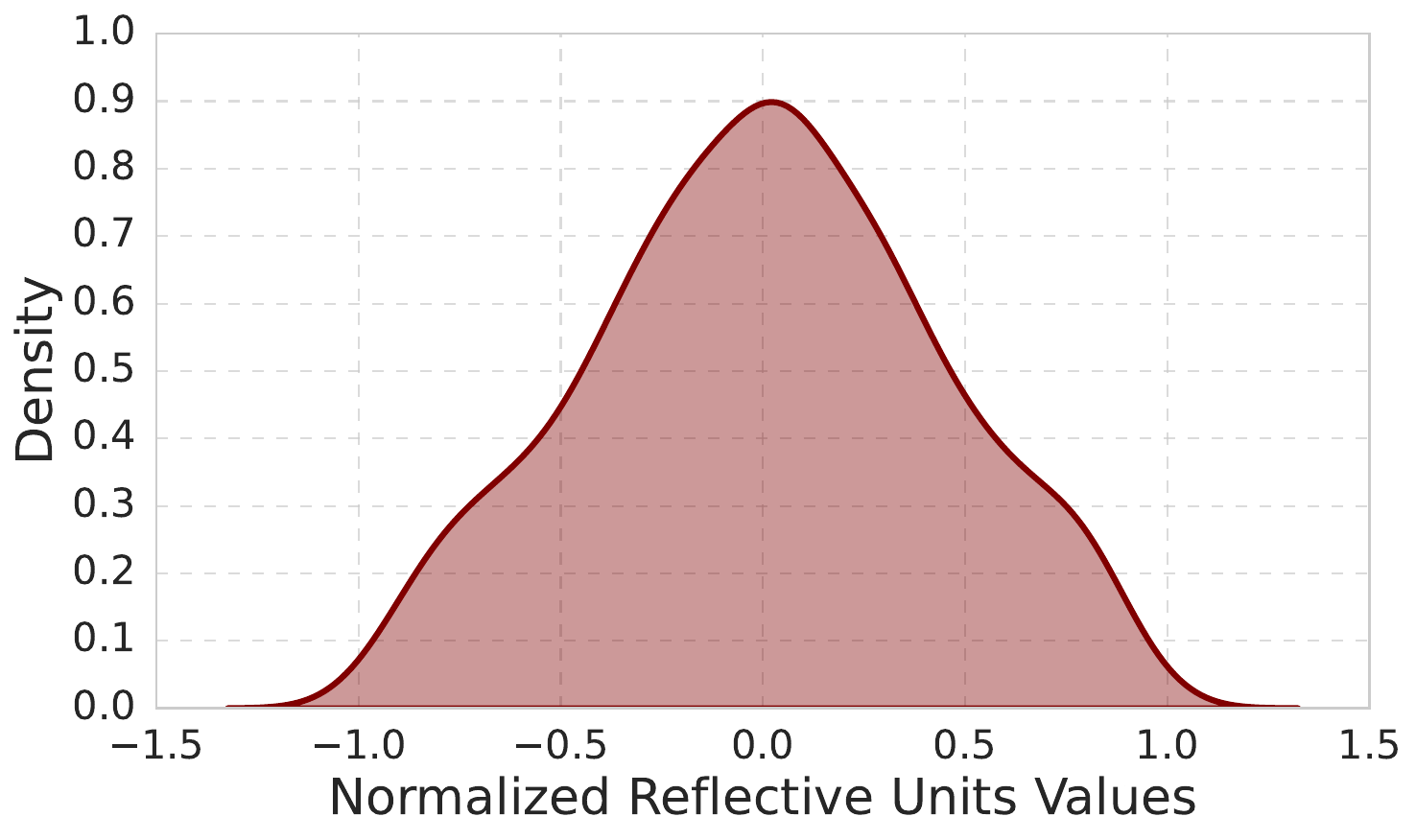}
        \caption{\texttt{GSM8K}}
        \label{fig:feature_distribution_gsm8k}
    \end{subfigure}
     \begin{subfigure}[b]{0.48\textwidth}
        \centering
        \includegraphics[width=\textwidth]{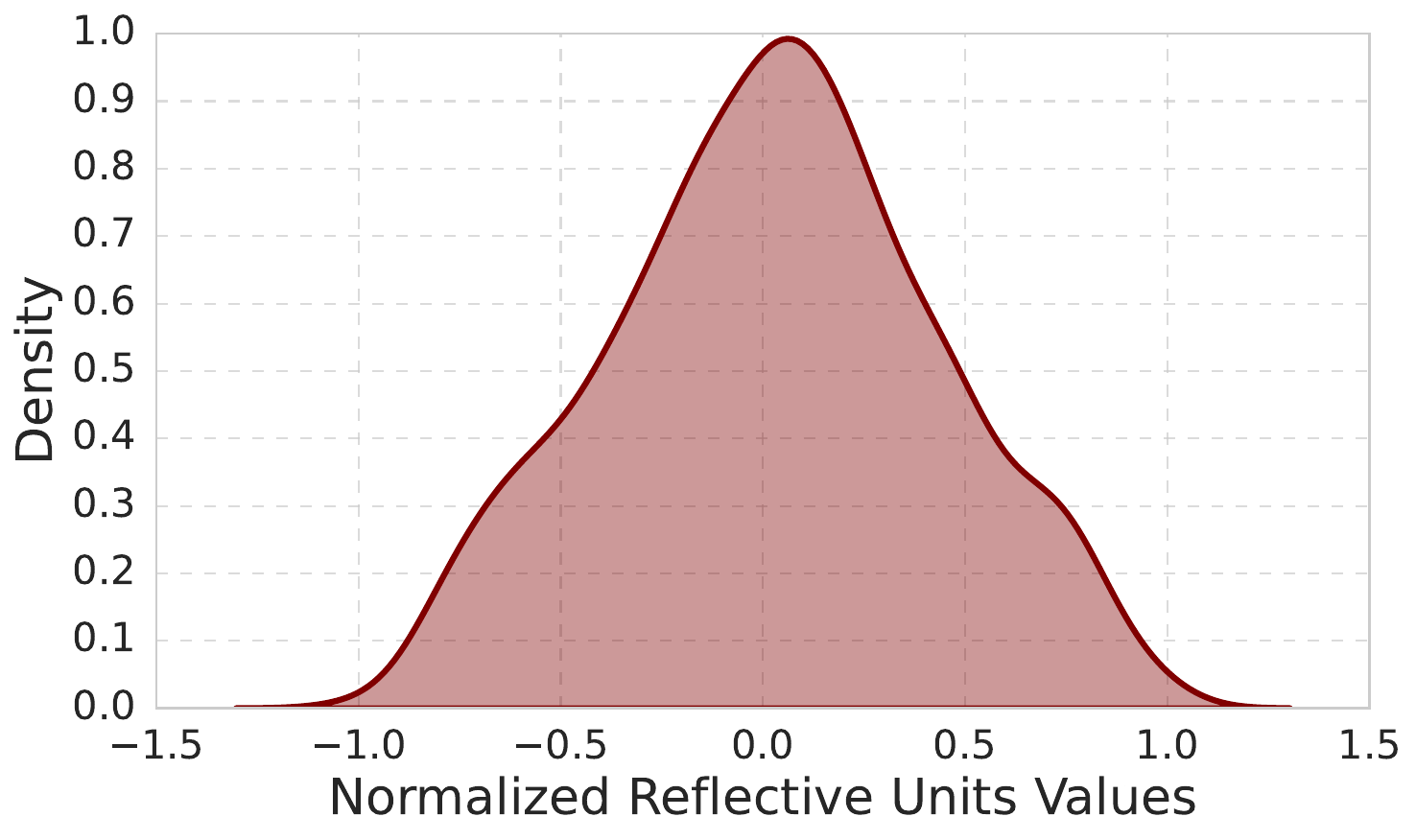}
        \caption{\texttt{ECID}}
        \label{fig:feature_distribution_ecid}
    \end{subfigure}
    \caption{Visualization of feature distributions for reflective units in the meta-reflection codebook.}
    \label{fig:feature_distribution_appendix}
\end{figure*}

Additionally, we visualize the feature distributions of reflective units in the meta-reflection codebook. Each reflective unit is first reduced to one dimension through dimensionality reduction and subsequently normalized. As shown in Figure~\ref{fig:feature_distribution_appendix}, the results demonstrate diverse distributions across reflective units, indicating their ability to capture varied semantic information without feature space collapse~\cite{jingunderstanding}.

\end{document}